\documentclass[format=acmsmall,screen]{acmart}

\usepackage{acronym}
\usepackage[inline]{enumitem}
\usepackage[normalem]{ulem}
\usepackage[skip=0pt]{caption} 
\usepackage{graphicx}

\AtBeginDocument{%
  \providecommand\BibTeX{{%
    \normalfont B\kern-0.5em{\scshape i\kern-0.25em b}\kern-0.8em\TeX}}}

\newcommand{\changed}[1]{\textcolor{black}{#1}}

\newcommand{\sww}[1]{\textcolor{black}{#1}}

\newcommand{\header}[1]{\vspace*{1mm}\noindent\textbf{#1.}}

\newcommand{\code}[1]{{\ttfamily#1}}

\acrodef{USS}{user satisfaction simulation}
\acrodef{MUSS}{metaphorical user satisfaction simulation}
\acrodef{TDS}{task-oriented dialogue system}

\newtheoremstyle{mydef}
{2ex}
{2ex}
{\itshape}
{}
{\scshape}
{: }
{0.5em}
{}
\theoremstyle{mydef}
\newtheorem{mydef}{Definition}

\copyrightyear{2023}
\acmYear{2023}
\setcopyright{acmlicensed}
\acmDOI{}
\acmISBN{}
\acmJournal{TOIS}
\acmVolume{1}
\acmNumber{1}
\acmArticle{1}
\acmDOI{}

\author{Weiwei Sun}
\affiliation{%
  \institution{Shandong University}
  \city{Qingdao}
  \country{China}}
\email{sunnweiwei@gmail.com}

\author{Shuyu Guo}
\affiliation{%
  \institution{Shandong University}
  \city{Qingdao}
  \country{China}}
\email{guoshuyu225@gmail.com}

\author{Shuo Zhang}
\affiliation{%
  \institution{Bloomberg}
  \city{London}
  \country{United Kingdom}}
\email{szhang611@bloomberg.com}

\author{Pengjie Ren}
\affiliation{%
  \institution{Shandong University}
  \city{Qingdao}
  \country{China}}
\email{renpengjie@sdu.edu.cn}

\author{Zhumin Chen}
\affiliation{%
  \institution{Shandong University}
  \city{Qingdao}
  \country{China}}
\email{chenzhumin@sdu.edu.cn}

\author{Maarten de Rijke}
\affiliation{%
  \institution{University of Amsterdam}
  \city{Amsterdam}
  \country{The Netherlands}}
\email{m.derijke@uva.nl}

\author{Zhaochun Ren}
\affiliation{%
  \institution{Shandong University}
  \city{Qingdao}
  \country{China}}
\email{zhaochun.ren@sdu.edu.cn}
\authornote{Corresponding author.}

\renewcommand{\shortauthors}{Sun et al.}

\acrodef{GBS}{Generalized Binary Search}

\begin{document}

\title[Metaphorical User Simulators for Evaluating Task-oriented Dialogue Systems]{Metaphorical User Simulators for Evaluating Task-oriented Dialogue Systems}



\begin{abstract}
Task-oriented dialogue systems (TDSs) are assessed mainly in an offline setting or through human evaluation.
The evaluation is often limited to single-turn or is very time-intensive. 
As an alternative, user simulators that mimic user behavior allow us to consider a broad set of user goals to generate human-like conversations for simulated evaluation.
Employing existing user simulators to evaluate TDSs is challenging as user simulators are primarily designed to optimize dialogue policies for TDSs and have limited evaluation capabilities. 
Moreover, the evaluation of user simulators is an open challenge. 

In this work, we propose a metaphorical user simulator for end-to-end TDS evaluation, 
where we define a simulator to be metaphorical if it simulates user's analogical thinking in interactions with systems.
We also propose a tester-based evaluation framework to generate variants, i.e., dialogue systems with different capabilities.
Our user simulator constructs a metaphorical user model that assists the simulator in reasoning by referring to prior knowledge when encountering new items.
We estimate the quality of simulators by checking the simulated interactions between simulators and variants. 
Our experiments are conducted using three TDS datasets. 
The proposed user simulator demonstrates better consistency with manual evaluation than an 
agenda-based simulator and a seq2seq model on three datasets; 
our tester framework demonstrates efficiency and has been tested on multiple tasks, such as conversational recommendation and e-commerce dialogues.

\end{abstract}


\maketitle

\section{Introduction}
\label{sec:intro}

\Acp{TDS} assist users in solving a specific task during a conversation~\citep{Young2013POMDPBasedSS}. 
They are being considered in a growing number of information retrieval scenarios, such as conversational information seeking, conversational Q\&A, and conversational recommendation~\citep{Gao2021AdvancesAC,Reddy2019CoQAAC,Ren2021WizardOS}.
In a \ac{TDS}, dialogues follow a clearly defined structure that builds on domain knowledge relevant for the task(s) at hand. 
The evaluation of \acp{TDS} is a crucial part of the development process.
Recent studies on evaluating \acp{TDS} are either through offline evaluation or human evaluation~\cite{Lamel2000TheLA,Jurccek2011RealUE}.
Offline evaluation evaluates a dialogue system based on test sets, whereas human evaluation reflects the overall performance of the agent through in-field experiments~\cite{Black2011SpokenDC,Lamel2000TheLA} or crowd-sourcing~\cite{Jurccek2011RealUE}.
However, offline evaluation is often limited to single turn assessments, while human evaluation is intrusive, time-intensive, and does not scale~\citep{Deriu2020SurveyOE}.

\header{User simulations}
User simulation can potentially mitigate the above issues via simulated interactions, and may thus be a viable choice for large-scale automatic evaluation for \ac{TDS}~\cite{Deriu2020SurveyOE}.
The structured nature of task-oriented dialogues allows us to build user simulators as we can exhaustively enumerate user goals to generate human-like conversations for simulated evaluation~\citep{Zhang2020RecentAA}.
Recent studies have employed user simulation for evaluating conversational information access tasks, including complex information seeking~\citep{Labhishetty2021AnEO}, conversational item recommendations~\citep{Zhang2020EvaluatingCR}, and multi-step task completion~\citep{Sun2021SimulatingUS}.
User simulators have also been successfully applied as a reinforcement learning environment for dialogue policy optimization~\citep{Shi2019HowTB}.
However, user simulators as evaluation methods for \ac{TDS}s are still under-explored~\citep{Balog2021Sim4IRTS}.

We identify the following challenges when employing user simulators to evaluate \Acp{TDS}:
\begin{enumerate*}[label=(\roman*)]
\item First,  employing existing simulators for evaluating \ac{TDS}s suffers from limited realism and evaluation capability~\citep{Balog2021Sim4IRTS}.
E.g., commonly used agenda-based simulators~\cite{Schatzmann2007AgendaBasedUS} rely on human-curated rules that are domain-specific; thus, their responses are far from human-like~\citep{Zhang2020EvaluatingCR}.
End-to-end simulators~\cite{Papangelis2019CollaborativeMD,Tseng2021TransferableDS} have been proposed to improve the domain generalizability of simulators. Still, their evaluation capabilities are limited at the semantic
level (e.g., if the system provides a novel entity)~\citep{Tseng2021TransferableDS}.
\item Besides, there is a lack of automatic methods to assess user simulators' realism and evaluation capabilities.
Most existing methods rely on manual evaluation, which is costly and hard to reproduce.
Thus, the evaluation of user simulators is still an open challenge~\citep{Pietquin2012ASO}.
\end{enumerate*}

\header{Our proposal}
In the face of the challenges identified above, we aim to provide a solution for improving the realism and transferability of user simulators and thus serve as an evaluation method for \acp{TDS}.
Our solution includes a simulator based on metaphorical user modeling for end-to-end \ac{TDS} evaluation and an automated evaluation method based on testers.

Specifically, we propose a metaphorical user simulator (MetaSim) that improves the realism of the conversation strategies by simulating the user's analogical thinking.
MetaSim is metaphorical because it uses historical strategies as metaphors for an ongoing dialogue: it assists the simulator in the reasoning process by referring to historical dialogue strategies when encountering new items.
MetaSim comprises:
\begin{enumerate*}[label=(\roman*)]
\item a preference module that initiates diverse user preferences and updates them during conversations; 
\item a natural language understanding (NLU) module that tracks the dialogue state; 
\item a metaphor module that retrieves similar dialogue strategies related to the current dialogue state; 
\item a policy module that simulates user satisfaction and predicts the user action according to context and retrieved strategies; and 
\item a natural language generation (NLG) module that generates a response in natural language. 
\end{enumerate*}

To evaluate user simulators, we construct a tester-based comparative evaluation framework inspired by the paradigm of evaluating interactive IR systems~\citep{Labhishetty2021AnEO} and conversational recommenders~\citep{Zhang2020EvaluatingCR}. 
A tester defines how to make variants for a dialogue system with distinguishable performance and how to evaluate user simulators by interacting with them.
We leverage a base dialogue system based on SOLOIST~\citep{Peng2020SOLOISTFT}. 
Its variants, defined by testers, are constructed by configuring parameters of the context history, retrieval or recommender methods, and dialogue domains.
We evaluate the simulators by checking if they can rank the system variants in the same way as humans and respond to new items in a human-like way.
A well-built tester can repeatedly evaluate the evaluation capabilities of simulators without the need for manual testing.

\header{Experiments}
We conduct experiments on three benchmark data\-sets, MultiWOZ~\citep{Eric2020MultiWOZ2A}, ReDial~\citep{Li2018TowardsDC}, and JDDC~\citep{Chen2020TheJC}.
We implement MetaSim based on T5~\citep{Raffel2020ExploringTL}, a pre-trained transformer model, and adopt a unified data format that allows it to be generalized to multiple tasks.
We optimize each module on the dataset and connect them during testing.
We evaluate MetaSim by checking the naturalness of the generated responses by user simulators, the quality of interactions, and consistency between user simulators and humans when comparing the variants of \Acp{TDS}.
Our experiments show that MetaSim consistently outperforms previous comparable simulators in terms of evaluation consistency and human-likeness.

Our experiments demonstrate that 
\begin{enumerate*}[label=(\roman*)]
\item metaphorical user simulators (MetaSim) can generate human-like dialogues and achieve evaluation results that are consistent with human expectations;
\item a tester-based framework is a valid tool for automatic assessment of simulator evaluation capabilities; and
\item our method generalizes well across domains.
\end{enumerate*}

\header{Contributions}
The contributions of this paper are:
\begin{itemize}
    \item We introduce a simulation approach for evaluating task-oriented dialogue systems. In particular, we propose a 
    metaphorical user simulator (MetaSim) that 
    leverages the dialogue records to improve the dialogue generation ability.
    \item We introduce a tester-based framework for evaluating user simulators; 
    \item We validate the proposed method through automatic and human experiments on three datasets; and
    \item We release the dataset and code of user simulators at \url{http://github.com/sunnweiwei/MetaSim} and the code of tester-based framework at \url{https://github.com/Superbooming/simtester}.
\end{itemize}
\section{Related Work}
\label{sec:related_work}

\subsection{Task-oriented dialogue systems}

Task-oriented dialogue systems (TDSs) aim to assist users in completing tasks through conversations~\citep{Eric2020MultiWOZ2A}.
TDSs are developed either via module-based or end-to-end approaches\changed{, differing in whether they manage the sub-steps of dialogue generation with multiple independent modules or not.} \citep{Jannach2021ASO}.
Research on the former boils down to advancing the components of \acp{TDS}, including natural language understanding (NLU), dialogue state tracking (DST), dialogue policy learning (DPL), and response generation (RG) \cite{Mrksic2015MultidomainDS,Mrksic2017NeuralBT,RojasBarahona2017ANE}.
The latter generates responses relying on end-to-end neural techniques in dialogue generation, such as the pre-trained language models, which aim to reduce the efforts in component-specific designs~\citep{Jin2018ExplicitST,Lei2018SequicityST,Vinyals2015ANC}. 
For example, 
\citet{Lei2018SequicityST} propose the \emph{belief spans} to track dialogue believes for TDSs, which allows an end-to-end seq2seq modeling. 
\citet{Liang2020MOSSED} introduce an end-to-end trainable framework for TDSs that aggregates supervision from various intermediate dialog system modules.

Recently, pre-trained language models have boosted the end-to-end dialogue modeling, e.g., GPT-2 for all sub-tasks involved in the task-oriented dialogue system~\citep{HosseiniAsl2020ASL, Yang2021UBARTF}. 
Pre-trained models based on heterogeneous dialogue corpus~\citep{Peng2020SOLOISTFT, He2021GALAXYAG} or leveraging multi-task learning~\citep{Su2021MultiTaskPF} have demonstrated promising performance in scenarios where in-domain annotated data is scarce~\citep{Madotto2020LanguageMA, Lin2021LeveragingSD}.
\acp{TDS} have also been considered in an increasing number of interactive information retrieval scenarios, such as conversational recommendation~\citep{Gao2021AdvancesAC,Li2018TowardsDC}, conversational information seeking~\citep{Ren2021WizardOS}, and conversational question answering~\citep{Reddy2019CoQAAC}.
For example,
question-based user preference elicitation~\citep{Gao2021AdvancesAC} has been proposed to ask about items~\citep{Zou2020NeuralIC} or attributes~\citep{Lei2020EstimationActionReflectionTD},
and multi-turn conversational recommendation strategies have been explored by \citep{Sun2018ConversationalRS,Zhang2018TowardsCS}.
\citet{Li2018TowardsDC} construct REDIAL, which is an annotated dataset of conversational recommendation dialogues about movies.
\citet{Ren2021WizardOS} propose a module-based framework for conversational information seeking.

\subsection{Evaluation of dialogue systems}
The typical evaluation methods of dialogue systems include automatic evaluation and human evaluation \citep{Deriu2020SurveyOE}.
Commonly used automatic evaluation methods include component-level evaluation and task-level evaluation. 
Component-level evaluation is concerned with the assessment of each component.
For example, the DST module is conventionally evaluated by joint goal accuracy (i.e., the average accuracy of predicting all slot assignments for a turn correctly), and the accuracy of action prediction is used to evaluate the dialogue policy learning module~\citep{Eric2020MultiWOZ2A}. 
\changed{To evaluate the item recommendation capability, metrics like recall and NDCG are used to measure the accuracy of recommended items per turn~\citep{Ren2022VariationalRA}.}
As for the evaluation of response evaluation, BLEU~\citep{Papineni2002BleuAM} is employed to measure the word overlap between generated response and ground truth, which draws inspiration from the evaluation of machine translation; Perplexity and Distinct are also used to evaluate the fluency and diversity of text generation, respectively~\citep{Deriu2020SurveyOE}.

Task-level evaluation evaluates the overall performance of the dialogue system; commonly used metrics include task success rate, inform success rate, and average turn.
\changed{Specifically, task success rate validates whether the final recommended items meet the user's needs; }
inform success rate validates if the system answers the user's question; 
and average turn evaluates the efficiency of dialogue systems~\citep{Eric2020MultiWOZ2A,Gao2021AdvancesAC,Pathak2022ASO}.

The above methods based on a static test set are limited to a single turn and do not inform us about the overall usefulness of the system in interaction~\cite{Zhang2020EvaluatingCR}. Human evaluation can address this drawback \citep{Jurccek2011RealUE}, and popular human evaluation metrics include Satisfaction, Fluency, Coherence, Task success, Engagingness, etc.
However, human evaluation is intrusive, time-intensive, and does not scale~\citep{Smith2022HumanEO}. 
Employing simulation-based evaluation can tackle the above issues and be a viable choice for large-scale automatic evaluation \cite{Deriu2020SurveyOE}.

\subsection{User simulation}
User simulators are designed to simulate the user's behavior, which can be used either as an environment to train a reinforcement learning-based system or to evaluate a functioning system to find weaknesses (or assess) dialogue quality and potentially replace human evaluation~\cite{Deriu2020SurveyOE}.
For the sake of evaluating spoken dialogue systems, \citet{Eckert1997UserMF} propose the first statistical user simulator.
Using a Markov model~\citep{Georgila2005LearningUS}, \citet{Cuayhuitl2005HumancomputerDS} propose a hidden Markov model for the same purpose. 
In later work, the agenda-based user simulator~\citep{Schatzmann2007AgendaBasedUS} has been widely accepted as it elegantly  represents the user state as a stack of necessary user actions.

User simulation is not foreign to information retrieval evaluation; its importance has been confirmed in the Sim4IR workshop at SIGIR 2021~\citep{Balog2021Sim4IRTS}. 
Different methodologies have been proposed for building simulators, e.g., employing a Bayesian procedure~\citep{Carterette2011SimulatingSU}, cognitive state for interactive information retrieval~\citep{Maxwell2016AgentsSU}, and for different tasks like search sessions~\citep{Zhang2017InformationRE}, online news recommendation~\citep{Bountouridis2019SIRENAS}, conversational recommender systems via an agenda-based user simulator~\citep{Zhang2020EvaluatingCR}, biases present in the logged data~\citep{Huang2020KeepingDB}. 
In addition,
\citet{Shi2019HowTB} investigate the design of user simulators as an reinforcement learning environment. 
\citet{Tseng2021TransferableDS} introduce a reinforcement learning approach based on end-to-end modeling.

Existing simulators mechanically inform the system of the slot, which limits the realism and the evaluation capability~\citep{Sun2021SimulatingUS}.
Thus, the main challenge of employing simulation is to build a realistic user simulator that can mimic natural user behavior to a realistic extent, i.e., ``human-likeness.''
Humans have complicated mental activities and display specific behaviors when interacting with  machines~\citep{Edlund2008TowardsHS}.
Existing studies have shown that improving human-likeness of either simulators~\citep{Zhang2020EvaluatingCR,Sun2021SimulatingUS,Zhang:2022:SIGIR} or dialogue systems~\citep{Edlund2008TowardsHS} calls for detailed modeling of \emph{human behavior} in dialogues.
Studies show that users build ``mental models'' in interacting with a machine, and that such mental models are formed through analogical thinking~\citep{Balog2021ConversationalAF,Jones2011MentalMA}. 
We aim to prototype an initial mental model in this work, namely the metaphorical user simulator.
A metaphorical user simulator has a metaphor model that
refers to historical dialogue strategies in the reasoning process.
This metaphor model resembles semi-parametric methods in NLP~\citep{Das2022Semiparametric}, a class of methods that assist a series of knowledge-intensive language tasks (e.g., question answering, fact checking~\citep{Petroni2021KILTAB}) by retrieving external knowledge.
Compared with previous semi-parametric methods in NLP,
our proposed metaphorical model further improves the model's capability in user action prediction on the task-oriented dialogue and considers the contextual information about the retrieved content (i.e., the multi-turn dialogue context).

\subsection{Evaluation of user simulators}
The evaluation of user simulators is an open challenge \cite{Pietquin2012ASO}.
Existing evaluation methods of user simulators include the following approaches:
\begin{enumerate*}[label=(\roman*)]
\item Simulators are treated as dialogue systems and then evaluated similarly through text generation metrics (e.g., BLEU, PPL), component-level metrics (e.g., action prediction accuracy), and task metrics (e.g., task success rate)~\citep{Nekvinda2021ShadesOB}.

\item Alternatively, assessments are performed based on the realism of simulated dialogues~\citep{Zhang2020EvaluatingCR}. There, descriptive statistics (e.g., the number of turns, words, slots, actions) of machine-machine dialogues data are compared to data from human-human dialogues, or use humans are asked to perform a Turing test.

\item In addition, human evaluation is a critical component that assesses the consistency between humans and simulators.
\changed{Typical metrics include scoring-based methods (e.g., SM-Turn, SM-Dialog) and comparison-based methods (e.g., PW-Turn, PW-Dialog)}~\citep{Smith2022HumanEO}.

\item Lastly, some work evaluates simulators by checking the performance of dialogue systems trained with the simulators by reinforcement learning~\cite{Tseng2021TransferableDS,Shi2019HowTB}.
\end{enumerate*}

To directly assess the evaluation capabilities of a simulator,
recently, a tester-based evaluation framework has been used in interactive search \citep{Labhishetty2021AnEO}, which has shown good consistency with human expectations while costing less and being more stable than human evaluation. 
Our work differs from \citep{Labhishetty2021AnEO} in the following aspects.
Testers in interactive search in~\citep{Labhishetty2021AnEO} are constructed by configuring the retrieval methods, changing information needs, or using different document collections. 
Compared to these testers in interactive search, testers in \acp{TDS} require a substantial change in design as dialogue systems involve additional components, such as NLU, NLG, state tracking, etc.
We focus more on these characteristics of \acp{TDS}, e.g., tracking user intent in multi-turn conversations and ablations on an embedding-based recommendation module when designing testers. 
Specifically, we control the systems' ability to track a multi-turn conversation by limiting the number of available turns to system; the recommender is ablated by deleting features from item embeddings.
In addition, we establish a framework that can be used for multiple conversational tasks, e.g., movie recommendation, restaurant reservations. 
The proposed framework gets rid of human effort in dialogue evaluation and supports multiple conversational tasks, thus can facilitate reproduction and expansion of the evaluation of task-oriented dialogues.

\section{Problem Formulation}
\label{sec:task}

Most evaluation methods for \acp{TDS} follow the PARADISE~\citep{Walker1997PARADISEAF} framework and estimate user satisfaction based on dialogue cost and task success via automatic, simulated, or human evaluation.
Simulation-based evaluation methods efficiently evaluate the system with interactions, but their current capacity is too limited to compensate for human evaluation~\citep{Balog2021Sim4IRTS}.
Building a human-like user simulator does not mean that we are after a perfect mirror of human behavior, let alone the replacement of humans. 
The simulator should be good enough to be an assessor that correlates well with human assessment in certain aspects and potentially reduces the reliance on human effort in the loop of evaluation.

Our goal is to build an evaluation framework for comparing task-oriented dialogue systems via user simulation. 
This framework, first, should be capable of configuring \emph{testers} for a given base dialogue system, i.e., a \emph{base model}.
\begin{mydef}[Base Model]
The base model is a standard dialogue system that is configurable, thus leading to potentially different capabilities. 
\end{mydef}
\begin{mydef}[Tester]
A tester, $\mathbb{T}$, is an instance of a base system, which defines how to configure a base system, e.g., via the choices of the NLG model. 
\end{mydef}
\begin{mydef}[Variant]
A system \emph{variant}, $S_i$, is an instance of  $\mathbb{T}$.
\end{mydef}
\noindent
Given a tester system $\mathbb{T}$, any pair of its variants $S^{\mathbb{T}}_1$ and $S^{\mathbb{T}}_2$ should have a sensible difference in components, training data, and/or methods, and thus display distinguishable system performance.

Second, the user simulators can be configured to evaluate dialogue systems on certain aspects by interacting with the systems.
Suppose $S_1$ has a better ability in recommending new items than $S_2$, and the simulator is expected to measure this difference. 
The measures by the simulator should be consistent and stable with human assessments, and can be reproduced under this framework.

Lastly, this framework enables us to explore more human-like aspects to see how realistic simulators can be.  
In this study, we investigate the possibility of building a simulator that responds to new knowledge in a human-like fashion by exploring using a human metaphor. Here, \emph{metaphor} is the module where the user draws on relevant conversational strategies and knowledge in the interaction~\citep{Kaal2012MetaphorIC} and is found to be the primary element of the construction of a \textit{mental model}~\citep{Jones2011MentalMA}.
The simulation framework will be detailed in Sect.~\ref{sec:method}.

\section{Framework}
\label{sec:method}
We introduce a framework aimed at improving the realism and transferability of user simulators for end-to-end \ac{TDS} evaluation.
This framework includes a metaphorical user simulator (MetaSim) for end-to-end TDS evaluation and an automated evaluation method based on testers.
In this section, we first introduce MetaSim (Sect.~\ref{sec:metasim}), which improves the evaluation capability of the simulator by simulating user's analogical thinking.
Then, we introduce the tester-based evaluation framework for evaluating the evaluation capability of the simulators (Sect.~\ref{sec:tester-framework}).

\begin{figure*}[t]
 \centering
 \includegraphics[width=1\textwidth]{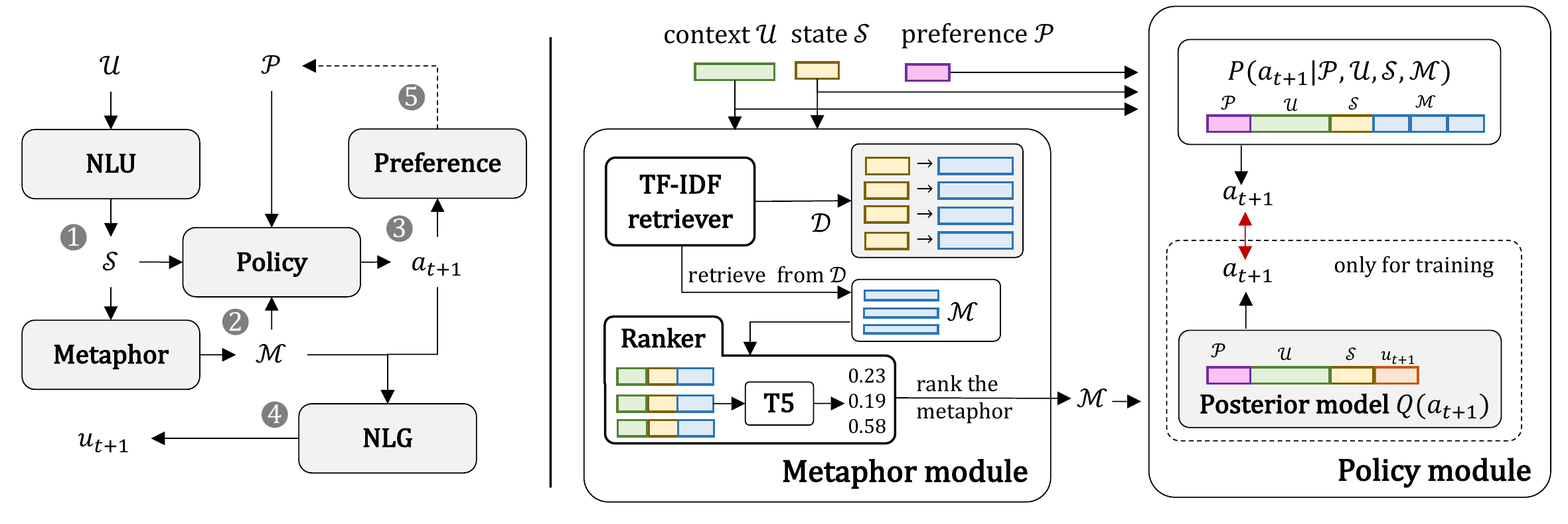}
 \caption{Architecture of the metaphorical user simulator (MetaSim). The left side shows the five modules of MetaSim (i.e., Preference, NLU, Metaphor, Policy, and NLG) and how they are connected. The right side details the Metaphor module and Policy module, where the Metaphor module retrieves $\mathcal{M}$ from $\mathcal{D}$, and the Policy module decides the next user action.}
 \label{fig:model}
\end{figure*}

\subsection{Metaphorical user simulators}
\label{sec:metasim}
We propose a metaphorical user simulator, \textbf{MetaSim}, that constructs a metaphorical user model to assist dialogue simulation by referring to prior knowledge.
The metaphorical user model simulates the user's analogical thinking in interacting with machines.
As shown in Figure~\ref{fig:model}, the simulator consists of five modules: a preference module, NLU, a metaphor module, a policy module, and \changed{an NLG module}. 

We introduce the modules one-by-one using the following notation: $\{\mathcal{P}$, $\mathcal{U}$, $\mathcal{S}$, $\mathcal{M}$, $a_{t+1}$, $u_{t+1}\}$, 
where
\changed{
$\mathcal{P}$ denotes user preferences that define the user requirements or goal of the conversation; 
$\mathcal{U}=\{u_{1}, r_{1}, \ldots, u_{t}, r_{t}\}$ is the dialogue context, in which $u_{i}$ denotes a user utterance at turn $i$, and $r_{i}$ denote the system's response at turn $t$; 
$\mathcal{S}=\{a_{1}, s_{1}, \ldots, a_{t}, s_{t}\}$ is the dialogue state while $a_{i}$ is the user's action at turn $i$, and $s_{i}$ is the dialogue belief state after system's response at turn $i$;
}
$\mathcal{M}=\{m_{1}, \ldots, m_{k}\}$ is the metaphor strategies
and each $m_{i}$ in $\mathcal{M}$ is a utterance;
$a_{t+1}$ is the next user action concatenated with user satisfaction, 
and $u_{t+1}$ is the next user utterance.
Besides, the metaphorical user simulator relies on a database $\mathcal{D}$ that contains dialogue records from which the model retrieves a metaphor $\mathcal{M}$, and we detail it in Sect.~\ref{sec:sec:metaphor}.

\subsubsection{Preference module}

The user preferences $\mathcal{P}$ define the user requirements or the goal of the conversation, and the preference module tells how $\mathcal{P}$ is initialized and updated during the conversations.
We represent the user preferences $\mathcal{P}$ as 
$$
\{(\text{slot}_{1},\text{value}_{1}), \ldots, (\text{slot}_{n},\text{value}_{n})\},
$$
where $\text{slot}_{i}$ denotes an attribute of user requirements and $\text{value}_{i}$ is its value.
\changed{
The initialization of $\mathcal{P}$ fills the initial user preferences before the conversation.
For example, the preference for hotel reservation ``\textit{want to book a cheap hotel where parking is available}'' is initialized as:
$\{(\text{hotel\_parking},``yes"),$
$(\text{hotel\_price},``cheap")\}$.
It can assist the model in answering the system's elicitation question at the beginning of the conversation, e.g., answer ``What price point would you like for the hotel?'' as ``cheap''.
The update of $\mathcal{P}$ tracks the user preference changes during the conversation.
For instance, the system may ask a question that is out of the scope of the initialized preference, such as ``do you need WIFI."
The preference module will add the newly generated user preferences, i.e., $(\text{WIFI\_needed}, ``yes'')$, to the initialized preferences when the answer is ``yes''. As a result, the model can give a consistent answer if the system asks a relevant question in future conversation turns.
}

\begin{figure*}[t]
 \centering
 \includegraphics[width=\textwidth]{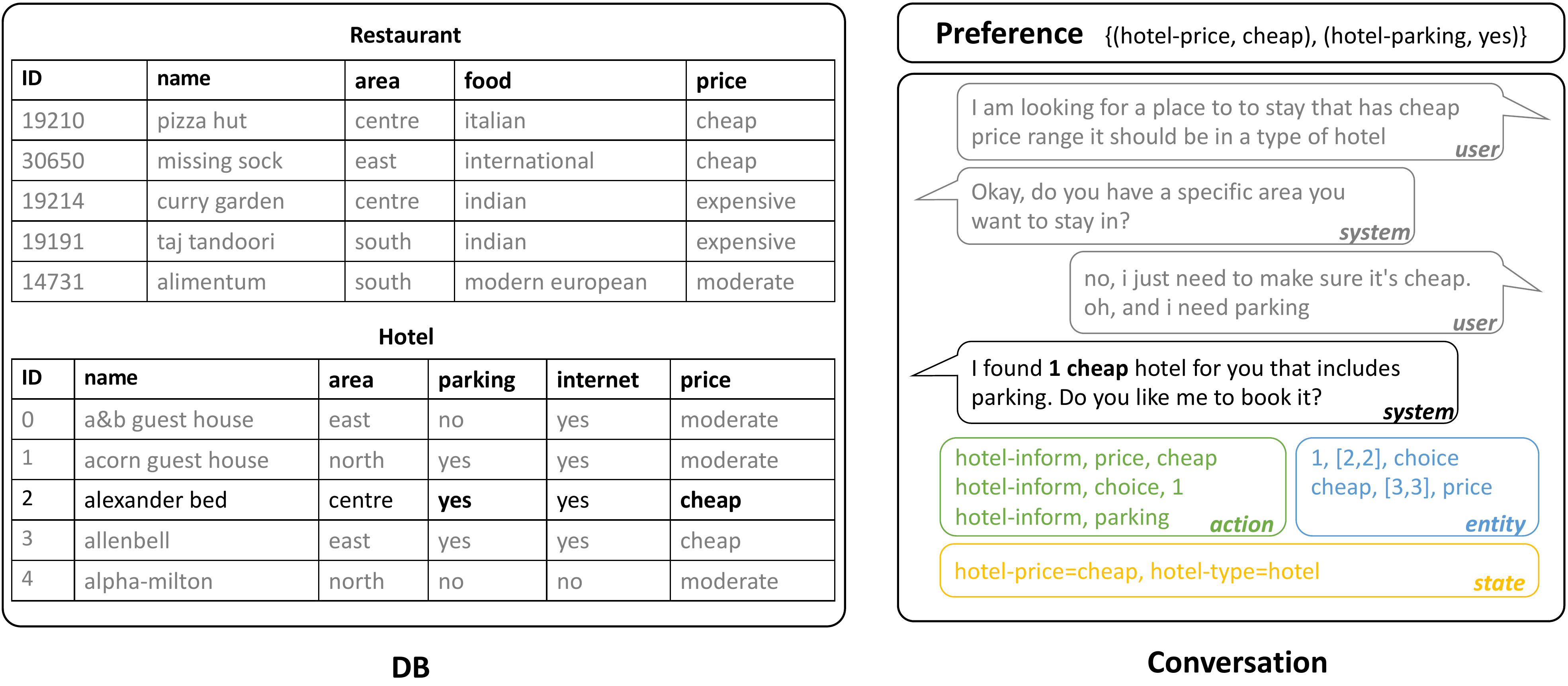}
 \caption{An example of an item database and conversation annotation in MultiWOZ 2.1.}
 \label{fig:db}
\end{figure*}

\header{Initialization of $\mathcal{P}$}
The $\mathcal{P}$ is initialized based on a \emph{item database}, namely $DB$.
As shown in Figure~\ref{fig:db}, the $DB$ consists of multiple tables, each table contains items of a specific domain (e.g., restaurant, hotel) with their attributes (e.g., name, area, food, etc.).
The initialization of $\mathcal{P}$ can be divided to three steps:
\begin{enumerate}
    \item \textbf{Domain sampling:} 
    In task-oriented dialogues, the user's preference can span multiple domains~\citep{Eric2020MultiWOZ2A}. For example, the user may want to find an attraction and then take a taxi there. We refer to a set of domain covered by a user's preferences as a \emph{domain combination}, where multiple domains are typically logically related and will constitute a joint goal. To sample the domain of the user's preference,
    we calculate the probability of each domain combination (e.g., $\{\text{hotel}\}$ and $\{\text{hotel}, \text{restaurant}\}$ denotes two different domain combinations) in the training set. 
    Then, we sample a domain combination based on the distribution of training set.
    In the example, the sampled domain combination is $\{\text{hotel}\}$.
    \item \textbf{Items sampling:} For each selected domain from \emph{Domain sampling} step, we sample one item from the table \changed{following} a uniform probability distribution.
    In the example, the ID of sampled item is $2$.
    \item \textbf{Attributes sampling:} We calculate the attribute number distribution using the training set, which defined as the number of slots used in the preference.
    Then, we sample $k$ attributes for each item follow a uniform probability distribution, and $k$ is determined follows the attribute number distribution.
    In the example, $k=2$ attributes are selected to be used in preference, that is parking=$``yes"$ and price=$``\mathit{cheap}"$.
\end{enumerate}

\header{Update of $\mathcal{P}$}
During the conversation, the preference module updates the preference $\mathcal{P}$ according to the following rules:
\begin{enumerate}
    \item If the user ``informs'' the system about a particular slot, then tag its value as ``Informed" in  $\mathcal{P}$. For example, for $\mathcal{P}=\{(\text{hotel\_parking},``yes"), (\text{hotel\_price},``\mathit{cheap}")\}$, if the user informs the system that the price is \emph{cheap}, then $\mathcal{P}$ is updated to  $\mathcal{P}=\{(\text{hotel\_parking},``yes")$, $(\text{hotel\_price},``\mathit{cheap} \mid \mathit{Informed}")\}$;
    \item If the system ``recommends'' an item, add it to $\mathcal{P}$. For example, if a newly recommended movie does not appear in $\mathcal{P}$, this module takes the movie as a slot and the user's favor towards it as value and adds it into $\mathcal{P}$.
\end{enumerate}

\subsubsection{Natural language understanding}
The NLU module keeps track of the states (i.e., the slots have been informed) of the dialogue.
We use \emph{actions} to represent the intent of an utterance \changed{(of either system or user)} in dialogue. 
We define the format of an action as follows:
$$
\{(\text{act}_{1},\text{slot}_{1},\text{value}_{1}), \ldots, (\text{act}_{n},\text{slot}_{n},\text{value}_{n})\},
$$
where each item $(\text{act}_{i},\text{slot}_{i},\text{value}_{i})$ includes:
\begin{enumerate*}[label=(\roman*)]
\item $\text{act}_{i}$, the name of action (e.g., \emph{inform}, \emph{bye}); 
\item $\text{slot}_{i}$, a slot, and 
\item $\text{value}_{i}$, its value.
\end{enumerate*}
Note that some actions that do not have slots or values, such as \emph{bye}; in such cases the slots and values are set to NULL.
Given user preferences $\mathcal{P}$ and dialogue context $\mathcal{U}$, 
the NLU module predicts the action $s_{t}$ of the last system response $r_{t}$. 
Then, we append $s_{t}$ to the historical actions\changed{, which are already predicted in previous turns. These steps form} the dialogue state $\mathcal{S}$.
Thus the dialogue state $\mathcal{S}$ records all the slots that have been informed from the system.
We optimize the NLU module using supervised learning and minimize the log-likelihood
$
\mathcal{L}_{NLU}=-\log(P(s_{t} \mid \mathcal{P}, \mathcal{U})).
$

\subsubsection{Metaphor module}
\label{sec:sec:metaphor}

The metaphor module is responsible for retrieving similar dialogue records from a database (terms as $\mathcal{D}$). The retrieved dialogue records are used to support the policy module.
To construct database $\mathcal{D}$, we index the dialogue state for each utterance in the training data and take the user utterance as the value.
Formally, $\mathcal{D}=\{(d_{1}^{k},d_{1}^{v}), \ldots, (d_{n}^{k},d_{n}^{v})\}$, 
where $n$ denotes that $\mathcal{D}$ has $n$ records, $d_{i}^{k}$ is the index of the $i$-th record, and $d_{i}^{v}$ is the value of the $i$-th record.

For any new dialogue, we retrieve $k$ records $\mathcal{M}=\{m_1,\ldots,m_k\}$ (we call $\mathcal{M}$ a \emph{metaphor}) from $\mathcal{D}$,  where each $m_i$ is a record retrieved from $\mathcal{D}$.

As illustrated in Figure~\ref{fig:model}, the retrieval follows a two-stage pipeline:
\begin{enumerate}
\item \textbf{Candidate generation}: For any record $(d_{i}^{k},d_{i}^{v})$, we retrieve $k$ candidate records from $\mathcal{D}$ using $\operatorname{TF-IDF}(\mathcal{S}, d_{i}^{k})$, which is the $\text{TF-IDF}$ score between the index of the record $d_{i}^{k}$ and the current dialogue state $\mathcal{S}$. 
These candidates form $\mathcal{M}$.

\item \textbf{Ranking}: We rank the candidates by a learnable \emph{ranker} for improving the retrieval accuracy. 
The \emph{ranker} is defined as $P(rel_i\mid m_i,\mathcal{U},\mathcal{S})$. Specifically, for an utterance $m_i$ in $\mathcal{M}$, we input the dialogue context $\mathcal{U}$ and dialogue state $\mathcal{S}$ to the \emph{ranker}, and estimate the relevance score \changed{$rel_i$} of them. 
In training, we set the relevance score of a ground-truth utterance to $1.0$ and the remaining utterances to $0.0$, and optimize the \emph{ranker} by maximizing the likelihood of the relevance score:
$$
\mathcal{L}_\mathit{Metaphor}=-\sum_{i=1}^{k}\log(P(rel_i \mid m_i,\mathcal{U},\mathcal{S})).
$$
In testing, we estimate the relevance scores for each utterance in $\mathcal{M}$ and rank them based on the scores.
\end{enumerate}

\subsubsection{Policy module}
\label{sec:sec:policy}
The policy module is responsible for predicting user satisfaction and the next user action.
We define it as
$P(a_{t+1}\mid \mathcal{P}, \mathcal{U}, \mathcal{S}, \mathcal{M})$.
The inputs of policy module are the user preferences $\mathcal{P}$, dialogue context $\mathcal{U}$, dialogue state $\mathcal{S}$, and metaphor $\mathcal{M}$; the output is the user action $a_{t+1}$. 
To predict the user satisfaction and action jointly, we make the user satisfaction a particular slot in the original action following~\citep{Sun2021SimulatingUS}.
The annotation of user action and satisfaction is usually incomplete or noisy~\citep{Han2021MultiWOZ2A,Sun2021SimulatingUS}, thus we introduce a \emph{posterior network} $Q(a_{t+1}\mid \mathcal{P}, \mathcal{U},\mathcal{S}, r_{t+1})$ to facilities the training of the policy module (as shown in Figure~\ref{fig:model}).
The \emph{posterior network} utilizes the ground-truth user utterance; thus, its predictions are more accurate than the policy module.
We first optimize the \emph{posterior network} on partially annotated data by a negative log-likelihood (NLL) objective:
$$
\mathcal{L}_\mathit{Policy}^{Q}=-\log(Q(a_{t+1})),
$$
and then optimize the policy module by learning from the \emph{posterior network}. The objective is defined as the Kullback–Leibler divergence between the $Q(a_{t+1})$ and $P(a_{t+1}\mid \mathcal{P}, \mathcal{U}, \mathcal{S}, \mathcal{M})$:
$$
\mathcal{L}_\mathit{Policy}^{P}=-Q(a_{t+1})\log\left(\frac{Q(a_{t+1})}{P(a_{t+1}\mid \mathcal{P}, \mathcal{U}, \mathcal{S}, \mathcal{M})}\right).
$$

\subsubsection{Natural language generation}
The NLG is responsible for generating natural language responses that incorporate system requests and individual preferences. 
We define it as:
$P(u_{t+1}\mid \mathcal{P}, \mathcal{U}, \mathcal{M}, a_{t+1})$, 
where the NLG module inputs the user preferences $\mathcal{P}$, dialogue context $\mathcal{U}$, metaphor $\mathcal{M}$ and next action $a_{t+1}$, generate the next utterance $u_{t+1}$ in a autoregressive fashion.
We optimize it by supervised learning; the training objective of the NLG module is defined as:
$$
\mathcal{L}_{NLG}=-\log(P(u_{t+1} \mid \mathcal{P}, \mathcal{U}, \mathcal{M}, a_{t+1})),
$$
where $u_{t+1}$ is the next user utterance, $\mathcal{P}$ is user preference, $a_{t+1}$ is the next user action, and $\mathcal{U}$ is the dialogue context.
    
\subsubsection{Inference}
\label{sec:sec:inference}
The operation of our model is shown in Figure~\ref{fig:model}.
We first draw a preference $\mathcal{P}$ from the preference module, that represents the initial user needs that lead the dialogue.
Then the simulator interacts with the system according to $\mathcal{P}$.
For each turn, with a dialogue history $U$, the model first predicts the system action $s_{t}$ by the NLU module and composes the dialogue state $\mathcal{S}$. 
\changed{Then the metaphor module retrieves $\mathcal{M}$ from training set dialogue records and ranks the record sentences using the ranker.}
After that, the policy module predicts the user action and satisfaction based on the metaphor and the dialogue context.  Then the NLG module converts the action to a natural language utterance $u_{t+1}$ and responds to the system. 
Finally, the preference module updates $\mathcal{P}$ by adding new items and removing informed slots.
The simulators generate ``[END]'' as an indicator to end the dialogue.

At the end of the dialogue, the simulator rates the system based on the simulated dialogue.
Previous studies~\citep{Sun2021SimulatingUS} show that human assessment of \ac{TDS} are mainly driven by the success of the task and the quality of system response. Thus we define the calibrated satisfaction by the average of \textsc{Success} (that equals $1$ if the system solves the user's problem and equals $0$ otherwise) and \textsc{Satisfaction} (that averages the turn-level user satisfaction during the dialogue).

\subsubsection{Implementation}
\label{sec:implementation}
\begin{figure}[t]
 \centering
 \includegraphics[width=0.76\columnwidth]{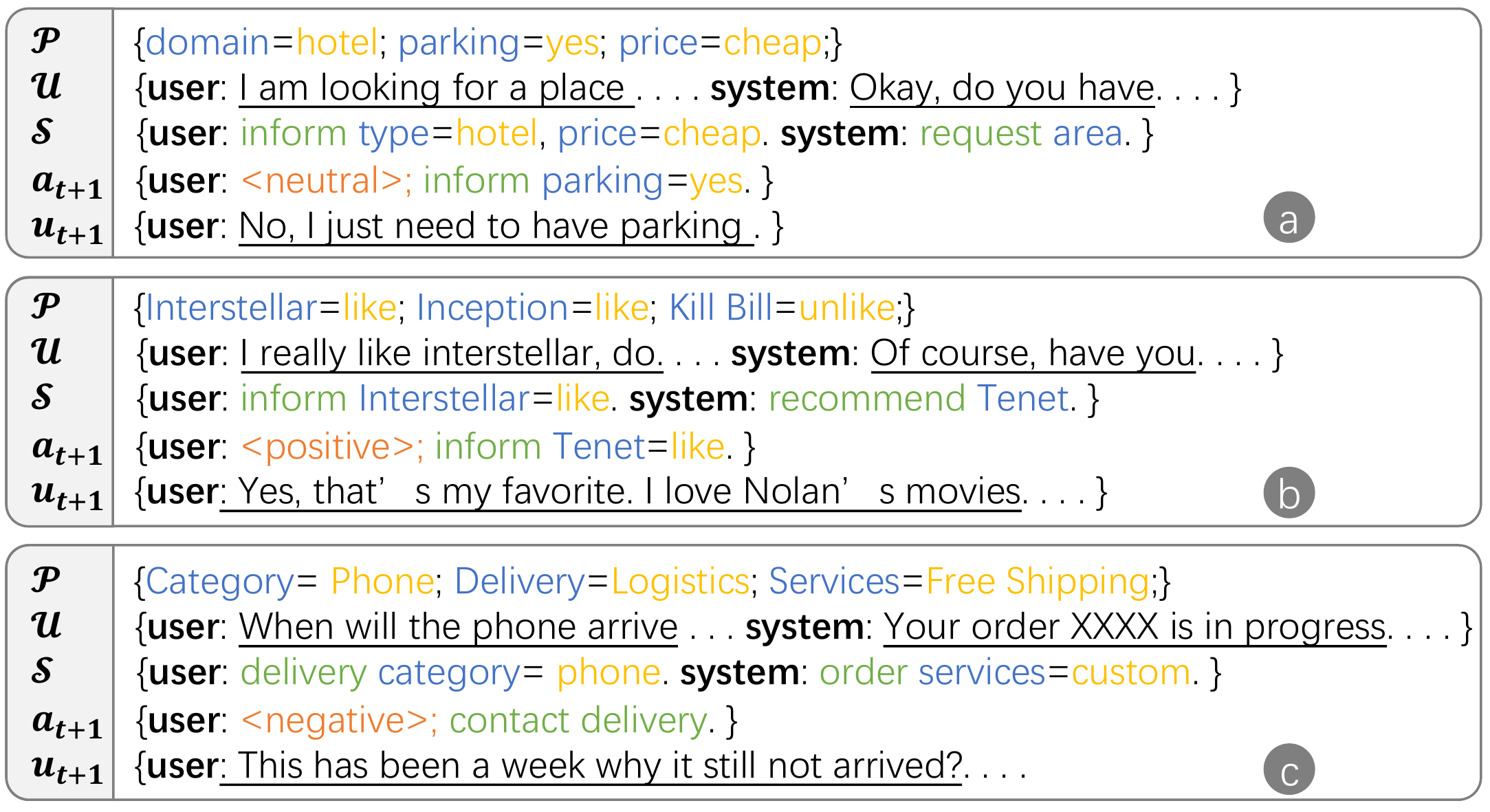}
 \caption{Examples of processed sequences. We highlight \textcolor[RGB]{79,113,190}{attributes},
 \textcolor[RGB]{246,194,67}{values},
 \textcolor[RGB]{126,171,85}{action},
 \textcolor[RGB]{223,130,68}{satisfaction}, 
 \uline{utterance}, and 
 \textbf{speaker}. (a) is in the \emph{hotel} domain from MultiWOZ, (b) is in the \emph{movie} domain from ReDial, and (c) is in the \emph{delivery} domain from JDDC.}
 \label{fig:examples}
\end{figure}

We use T5~\citep{Raffel2020ExploringTL} to instantiate each module. 
T5 is a sequence-to-sequence model with an encoder-decoder structure. We process the inputs and outputs of each module as a sequence. 
See Figure~\ref{fig:examples} for some examples of our sequence processing. 

For modules with multiple inputs, we concatenate them in order with a special token <SEP> and clip to the maximum input length of the model. 
Thus for each module, assuming the input text is $\mathcal{I}$ and the output text is $\mathcal{O}=\{o_{1}, \ldots, o_{m}\}$ with $m$ tokens, the probability is defined as:
$$
P(\mathcal{O}\mid\mathcal{I})=\sum_{i=1}^{m}P(o_{i}\mid\mathcal{O}_{<i}, \mathcal{I}).
$$

\noindent%
Taking the metaphor module as an example, we implement the ranker by T5, following techniques introduced by \citet{Nogueira2020DocumentRW}.
Recall that the metaphor module aims to retrieve similar dialogue records. Here, the learned model based on T5 first predicts the $\mathcal{O}=``yes"$ for positive records and $\mathcal{O}=``no"$ for negative records. The relevance score $r_i$ is then defined as $r_i=P(``yes" \mid m_i,\mathcal{U}, \mathcal{S})$.
%

\subsection{Tester-based evaluation framework}
\label{sec:tester-framework}
\label{sec:evaluation-framework}

\begin{figure}[t]
 \centering
 \includegraphics[width=0.7\columnwidth]{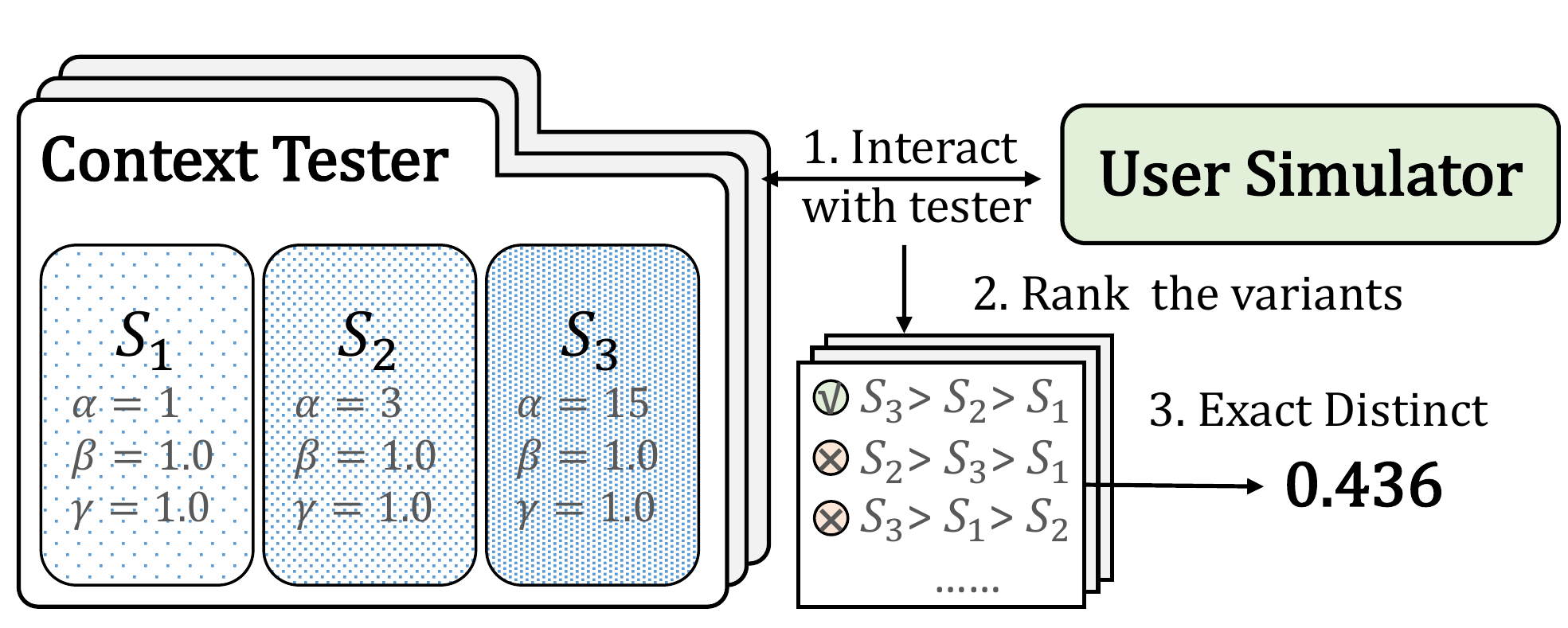}
 \caption{In tester-based evaluation, the simulator interacts with system variants in the tester, ranks them, and calculates the Exact Distinct score (i.e., to accurately distinguish between the three systems). }
 \label{fig:testers}
\end{figure}

Existing simulated evaluation methods for a \ac{TDS} typically rely on third-party systems~\citep{Zhang2020EvaluatingCR}.
This dependency suffers from the following issues: the limited choices of third-party systems and the difficulty of comparing them due to differences in training data and model architecture.
Drawing inspiration from \cite{Labhishetty2021AnEO}, we introduce a tester-based evaluation framework for evaluating both dialogue systems and user simulators, see Figure~\ref{fig:testers}.
The tester-based framework is constructed to evaluate the simulator automatically---the simulator talks to and evaluates the variants within a tester so that we can see whether a simulator can fairly compare the variants in a tester or not.
The tester-based framework has the following benefits in evaluating simulators: 
\begin{enumerate*}[label=(\roman*)]
\item the constructed system variants are well-calibrated and thus comparable to each other; 
\item a tester can evaluate any specific capability of a system, leading to a more comprehensive evaluation; and
\item a well-constructed tester does not require repeated human evaluation and can be reused to evaluate multiple simulators.
\end{enumerate*}

\subsubsection{Evaluating user simulators by testers}
Simulators are built for evaluating systems. Simulators also need to be assessed appropriately. Here, we discuss the latter case. The main idea to evaluate a simulator is to check whether the evaluation results by simulator upon variants are consistent with human expectations. For example, given a simulator $US$, and three variants $S_1$, $S_2$ and $S_3$, $US$ interacts with $S_1$, $S_2$ and $S_3$ separately and scores them based on the interactions. 
The performance scores are on certain aspects like recommendation accuracy and domain understanding. A comprehensive assessment requires various testers and variants (see Sect.~\ref{sec:sec:tester}). We compare these scores against human expectations. A better consistency score indicates a better simulator. 
    
We calculate the \textsc{ExactDistinct} (\textsc{ED} for short) score to measure the consistency between simulators' assessment scores and human expectations. \textsc{ExactDistinct} is defined as the proportion of examples that ranked the same as expectations by the human evaluator (i.e., $S_1<S_2<S_3$). An \textsc{ED} score of 0.5 means a rank accuracy of 50\%.

\subsubsection{Tester bootstrapping}
\label{sec:sec:tester}

We introduce the base system and three strategies to configure the testers.
We leverage a base model based on SOLOIST \cite{Peng2020SOLOISTFT}, which employs a sequence-to-sequence model for dialog state tracking and response generation. 
The testers enable us to configure the base model and make variants based on the needs. Specifically, we can bootstrap testers by changing the context in the dialogue history, exploring different choices for components like the recommender module, leveraging domain-specific training data, and supplying various amounts of supervision for training, to name a few.
Here, we introduce three strategies that we explore in this work; other options are left for future investigation: 

\begin{itemize}
\item \textbf{Context tester.}
    Context refers to the dialogue history. The amount of dialogue history that is used for training has an impact on the model performance~\citep{Gao2021AdvancesAC}. The Context tester, $\mathbb{T}_C$, makes variants by giving different context memory capabilities. We define a hyperparameter $\alpha$ indicating the number of sentences in the dialogue history. For example, $S_{\alpha=2}$ represents a model that keeps two sentences of dialogue history.
    
\item \textbf{Recommender tester.}
    The Recommender tester, $\mathbb{T}_R$, consists of systems with different retrieval capabilities. This tester demonstrates how to make variants by changing a system component. The recommender retrieves from an item database using a query that contains a set of keywords. We then control the capacity of the recommender system by removing the keywords in the query.
    We define a hyperparameter $\beta$ in the range 0 to 1. $S_{\beta=0.2}$ represents the model that keeps $0.2$ of features in query.

\item \textbf{Domain tester.}
    Domain tester, $\mathbb{T}_D$, makes variants by training modules that use different amounts of training data so that we can test the system's ability in a particular domain. We control it by removing training data from the training set. We define a hyperparameter $\gamma$ to indicate the number of dialogues used for training a dialogue system; $S_{\gamma=0.8}$ represents the model that uses $80\%$ of data.
\end{itemize}
\section{Experimental Setup}
\label{sec:exp}


Next, we seek to assess the proposed simulators  testers. 
To this end,
we first describe three datasets used in our experiments (see Sect. \ref{sec:sec:datasets}) 
and the implementation details of the user simulators and testers (see Sect. \ref{sub:sub:ief}).


\begin{table}[!t]
\setlength \tabcolsep{7pt}
\caption{Statistics of the training datasets. \changed{The top block gives the statistics of the original three datasets, and the bottom block lists the statistics of the USS dataset.}}
\label{table:data}
\begin{tabular}{l rrr}
\toprule

\textbf{Dataset} & \textbf{MultiWOZ} & \textbf{ReDial} & \textbf{JDDC} \\
\midrule
Language   & English & English & Chinese \\
Number of dialogues & 9,438 & 10,006 & 47,389 \\
Number of utterances & 147,176 & 142,502 & 633,826 \\
Avg number of turns  & 15.59 & 14.24 & 13.37 \\
Number of Action    & 13 & 4 & 19 \\
Number of Slots     & 15 & 6,925 & 23 \\
\midrule
\multicolumn{4}{l}{\textit{USS user satisfaction annotation}~\citep{Sun2021SimulatingUS}}\\
Number of dialogues & 1,000 & 1,000 & 3,300\\
Number of unsatisfied turns & 737 & 740 & 4,940\\
Number of fair turns &  11,141 & 9,623 & 45,005\\
Number of satisfied turns & 675 & 1,524 & 4,572\\
\bottomrule
\end{tabular}
\end{table}

\subsection{Datasets}
\label{sec:sec:datasets}

We conduct experiments on three datasets:
\begin{itemize}
    \item \textbf{MultiWOZ}~\citep{Eric2020MultiWOZ2A}: 
    MultiWOZ is a multi-domain task-oriented dialogue dataset spanning seven domains (i.e., Attraction, Hotel, Hospital, Police, Restaurant, Train, and Taxi). 
    It contains about 10K dialogues. Each dialogue can span up to three domains, and multiple domains might be mentioned in a single turn. MultiWOZ provides \emph{action}, \emph{entities} and \emph{dialogue state} annotations for each utterance.
    Figure~\ref{fig:db} shows an example.
    The \emph{action} provides the action name, and corresponding slot and value.
    The \emph{entities} provides the entity span, start and end position in text, and the corresponding slot name.
    The \emph{dialogue state} provides the (slot, value) pairs that the user has provided up to current turn.
    We use the MultiWOZ 2.1~\citep{Eric2020MultiWOZ2A} version, which fixes noisy state annotations across 40\% of the dialogue turns in previous versions.
    
    \item \textbf{ReDial}~\citep{Li2018TowardsDC}: 
    ReDial is conversational recommendation datasets; it consists of over 10K conversations, where users recommend movies to each other.
    We adopted this dataset and aligned it to task-oriented dialogue to verify the generalizability of the proposed framework.
    Following~\citep{Ma2021CRWalkerTG}, we identify entities in the dialogues by the Spacy toolkit\footnote{\url{https://spacy.io/}} and link the entities to a movie knowledge graph provided by DBpedia\citep{Auer2007DBpediaAN}. 
    In ReDial, the movie mentions and whether the user likes it or not is annotated. There is no action annotation in ReDial, so we constructed action as shown in Figure~\ref{fig:examples}.
    We then define a user's preferences as a list of movies the user has seen and whether they like them.
    
    \item \textbf{JDDC}~\citep{Chen2020TheJC}: JDDC is a large-scale, real-world Chinese e-commerce conversation corpus with over 1 million dialogues.
    We identify entities in the dialogues using the LTP toolkit\footnote{\url{http://www.ltp-cloud.com/}} and link the entities to an e-commerce knowledge graph provided by Aliyun.\footnote{\url{https://www.aliyun.com/}}
    We define the user's preference as the entities mentioned in the dialogues and design an action based on the informed entity (as shown in Figure~\ref{fig:examples}).
    Since the original dataset is large and excessive in topics, we sample the dialogues related to the ``delivery'' domain from JDDC, resulting in 47K dialogues.
\end{itemize}
We additionally leverage the USS~\citep{Sun2021SimulatingUS} dataset, which provides user satisfaction annotations (on a 5-point scale) both on the exchange level and for entire conversations for the three datasets listed above.
\begin{itemize}
    \item \textbf{USS}~\citep{Sun2021SimulatingUS}: \changed{
    The USS extends a subset of the above datasets with user satisfaction annotations. As per Sect.~\ref{sec:sec:policy}, we optimize the posterior policy network on partially annotated data and use the posterior network to label the satisfaction on the remaining unlabeled data for the training of the other modules. 
    The posterior network achieved 97.79\% accuracy and 88.72\% USR on the test set USS. We then leverage these predicted satisfaction labels.
    We modify the 5-point scale satisfaction annotation of USS (i.e., 1 = Very unsatisfied, 2 = Unsatisfied, 3 = Fair, 4 = Satisfied, 5 = Very satisfied) to a 3-point scale (1 = Unsatisfied, 2 = Fair, 3 = Satisfied).
    Since USS's satisfaction distribution is unbalanced (most turns express fair satisfaction), we up-sample the non-fair satisfaction turns by increasing the sampling probability of non-fair samples by 10 times.
}
\end{itemize}
Summary statics of our training datasets are in
Table~\ref{table:data}. 
\changed{The processed datasets are available at \url{http://github.com/sunnweiwei/MetaSim}.}

\begin{table*}[!t]
\centering
\caption{Test set evaluation results of user simulators. Dist. is short for \textsc{Distinct}. Boldface indicates the best results in the corresponding metric; significant improvements over the best baseline are marked with * (t-test, $p<0.05$).}
\label{table:sim-test}
{%
\begin{tabular}{l ccc ccc ccc}

\toprule

{\textbf{Datasets}} 
& \multicolumn{3}{c}{\textbf{MultiWOZ}} 
& \multicolumn{3}{c}{\textbf{ReDial}} 
& \multicolumn{3}{c}{\textbf{JDDC}}\\

\cmidrule(lr){2-4} \cmidrule(lr){5-7}  \cmidrule(lr){8-10}  

& F1     & Dist.   &  Slot  
& F1     & Dist.   &  Slot  
& F1     & Dist.   &  Slot  
\\ 

\midrule

Agenda~\citep{Schatzmann2007AgendaBasedUS}
& 27.25    &  18.18   & 27.73  
& 16.58    &  \textbf{24.36}   & 7.44  
& 13.18    &  22.42   & 8.9
\\

AgendaGen~\citep{Shi2019HowTB}
& 34.51    & 16.91    & 24.68
& 21.06    & 18.81    & 7.15 
& 17.32    & 25.36    & 10.53  
\\

SL+Template~\citep{Shi2019HowTB}
& 35.54    & \textbf{20.20}    &  42.02 
& 27.83    &  23.27   &  15.07   
& 12.91    &  25.28   &  40.55  
\\

Seq2seq~\citep{Raffel2020ExploringTL}
& 42.02    &  14.98   & 49.41  
& 29.75    &  16.97   & 11.28  
& 23.45    &  21.29   & 37.26  
\\

\midrule

\textbf{MetaSim}
& \textbf{44.13}\rlap{*}    & 17.29    & \textbf{57.65}\rlap{*}  
& \textbf{33.79}\rlap{*}    & 20.43    & \textbf{15.72}\rlap{*}    
& \textbf{23.85}    & \textbf{26.37}\rlap{*}    & \textbf{44.00}\rlap{*}    
\\

MetaSim-Metaphor
& 43.31    & 16.83    & 55.27   
& 32.59    & 18.94    & 15.21  
& 23.42    & 25.41    & 43.65 
\\

MetaSim-Policy
& 43.11    & 14.65    & 49.46
& 33.67    & 15.80    & 13.01  
& 22.35    & 20.46    & 39.55 
\\

MetaSim-Preference
& 33.84    &  12.71   & \phantom{1}9.76    
& 32.18    &  14.25   & \phantom{1}4.35    
& 18.13    &  19.76   & 22.32 \\
\bottomrule
\end{tabular}
}

\end{table*}

\begin{table*}[!t]
\centering
\setlength\tabcolsep{2.5pt}
\caption{Interaction and human evaluation results of user simulators. Int.\ is short for Interaction, Dist. is short for \textsc{Distinct}, Coh. is short for \textsc{Coherence}, and Hum. is short for \textsc{Human-like}. Boldface indicates the best results in the corresponding metric; significant improvements over the best baseline are marked with * (t-test, $p<0.05$).
}
\label{table:sim-human}
{%
\begin{tabular}{l  cccc cccc cc}

\toprule

{\textbf{Datasets}} 
& \multicolumn{4}{c}{\textbf{MultiWOZ}} 
& \multicolumn{4}{c}{\textbf{ReDial}} 
& \multicolumn{2}{c}{\textbf{JDDC}}\\

\cmidrule(lr){2-5} \cmidrule(lr){6-9}  \cmidrule(lr){10-11}  

{}
& \multicolumn{2}{c}{Inter.}
& \multicolumn{2}{c}{Human}
& \multicolumn{2}{c}{Inter.}
& \multicolumn{2}{c}{Human}
& \multicolumn{1}{c}{Inter.}
& \multicolumn{1}{c}{Human} 
\\

\cmidrule(lr){2-3} \cmidrule(lr){4-5} \cmidrule(lr){6-7} \cmidrule(lr){8-9}
& Success  & Dist.   & Coh.   & Hum.
& Success  & Dist.   & Coh.   & Hum.
& Dist.   & Hum.
\\ 

\midrule

Agenda~\citep{Schatzmann2007AgendaBasedUS}
& 12.19  & 14.15    & 1.96     &  1.84
& 4.84    & 22.97    & 2.45    &   1.95
&  29.37   & 1.81
\\

AgendaGen~\citep{Shi2019HowTB}
&  32.17   & 13.58    &  2.20   & 1.98   
&  6.78   & \textbf{23.04}    & 2.42    &   2.10 
&  31.49   &   1.97 
\\

SL+Template~\citep{Shi2019HowTB}
&  59.78   & 12.74    &  2.20   &   1.75
& 7.30   & 19.26    & 2.64    &    2.30
&  28.74   &     1.86
\\

Seq2seq~\citep{Raffel2020ExploringTL}
&  76.75   &  13.76   & 2.18    &   2.04
& 6.18   &  16.45   &  2.30   &  2.35  
& 23.29    &    2.22
\\

\midrule

\textbf{MetaSim}
& \textbf{80.25}\rlap{*}   & \textbf{15.65}    &   \textbf{2.31}  & \textbf{2.21}  
& \textbf{13.70}\rlap{*}    &  20.97   &  \textbf{2.70}   &  \textbf{2.41}  
& \textbf{34.81}\rlap{*}    &    \textbf{2.38}
\\

MetaSim-Metaphor
& 78.96    & 15.32    & 2.25    &   2.13
& 12.74    & 19.36    & 2.66    &    2.35
& 32.65    &  2.31 
\\

MetaSim-Policy
& 71.54    & 14.18    &  2.20   &  2.17
& 12.87    &  17.39   & 2.35    & 2.14
&  30.11   &    2.14
\\

MetaSim-Preference
&  \phantom{1}0.93   &  \phantom{1}5.20    & 1.98    &   1.72
& \phantom{1}3.43   & \phantom{1}5.09    & 2.05    &  2.05  
&  \phantom{1}0.09   &  1.77  
\\
\bottomrule
\end{tabular}
}

\end{table*}

\subsection{Implementations of the evaluation framework}
\label{sub:sub:ief}

\textbf{Implemented user simulators.}
We use T5-base \citep{Raffel2020ExploringTL} as the backbone of MetaSim for MultiWOZ and ReDial, and optimize the model by AdaFactor with a learning rate of 1e-3.
For the Chinese dataset JDDC, we use Mengzi-T5-base \cite{Zhang2021MengziTL} and use AdamW with a learning rate of 5e-4.
We set the batch size to 64, use a linear lr scheduler with 500 warm-up steps, and train up to 20 epochs.
We set the max length to 512 for MultiWOZ and 256 for ReDial and JDDC.
We perform greedy decoding for simulators.
Due to the characteristics of the data, we replaced the NLU module on the ReDial dataset with a rule-based model that identifies and links the entities in the system utterance to the knowledge graph~\citep{Auer2007DBpediaAN}, and serialize the linked subgraphs as system actions.

We additionally implement the following baselines on three datasets for comparing against MetaSim:
\label{sec:sec:baselines}
\begin{itemize}
    \item \textbf{Agenda model} \citep{Schatzmann2007AgendaBasedUS} constructs an agenda (that consist of a stack of actions) and pops the top action at each dialogue turn. 
    We build an agenda model by statistically measuring the distribution of action transitions in the training dataset.
    We retrieve the templates that maximize the override of the slots in action and choose the one with the highest TF-IDF score between the dialogue context. 
    
    \item \textbf{Agenda Gen model} \citep{Shi2019HowTB} uses the same agenda-based strategy as the Agenda model but generates responses using a generative model instead of retrieved templates. We use a T5-base as the generative model and input the dialogue history and the action drawn from the agenda.
    
    \item \textbf{SL+Template} \citep{Shi2019HowTB} predicts employing a retriever in the same way as the Agenda model but using the action predicted by a neural model instead.
    We use an NLU module and a Policy module to predict the next user action, retrieve a template from the training data, and fill the slot in the template as the response.
    \item \textbf{Seq2seq model} \citep{Raffel2020ExploringTL} generates response based on dialogue history and preference with a sequence-to-sequence model. 
    We process the data into sequence format and train a T5-base to perform text-to-text generation. Compared to our model, the Seq2seq model doesn't model the policy and metaphor of the user explicitly.
\end{itemize}
We also compare MetaSim with three modifications:
\begin{itemize}
    \item \textbf{MetaSim-Metaphor} removes the metaphor module and always sets $\mathcal{M}$ to ``\textit{none}.''
    \item \textbf{MetaSim-Policy} removes the policy module and always sets the user action $a_{t=1}$ to ``\textit{none}.''
    \item \textbf{MetaSim-Preference} removes the $\mathcal{P}$ and let the model response without constrain.
\end{itemize}

\header{Implemented testers}
\label{sec:sec:iml-tester}
We use SOLOIST~\citep{Peng2020SOLOISTFT} as the base dialogue system, which is a sequence-to-sequence model based on GPT2~\citep{Radford2019LanguageMA}.
We implement the base system with hyperparameters of $\{\alpha=15,\beta=1,\gamma=1\}$.
Then we set the system variants in context tester as $\mathbb{T}_C=\{S_{\alpha=3}, S_{\alpha=1}\}$; 
set the system variants in recommender tester as $\mathbb{T}_R=\{S_{\beta=0.4}, S_{\beta=0.1}\}$; and 
set the system variants in domain tester as $\mathbb{T}_D=\{S_{\gamma=0.1}, S_{\gamma=0.01}\}$.
We re-implement the SOLOIST model for ReDail and JDDC. 

We use the same data annotation and processing for system implementation as for the user simulator (examples are given in Figure~\ref{fig:examples}).
The dialogue system for MultiWOZ and ReDial additionally includes a recommendation module.
In MultiWOZ, we generate SQL statements based on belief state.
For example, the SQL statements of the belief state in Figure~\ref{fig:db} is \code{select * from Hotel where price=cheap and type=hotel}.
The retrieved results are then fill the results back into the responses by string matching following \citet{Peng2020SOLOISTFT}.
In ReDial, we implement a recommender as a classifier following~\citep{Li2018TowardsDC} (i.e., retrieve the movie by calculating the inner-product of the hidden state produced by the GPT model the movie embeddings).
Since the database used in the JDDC dataset is not publicly available~\citep{Chen2020TheJC}, 
we skip the recommendation module in JDDC and do not implement a recommender tester on JDDC.

We optimize the systems using AdamW with a learning rate of 1.5e-4, batch size 32. 
The systems are trained up to 20 epochs.
We set the max length to 512 for MultiWOZ and to 256 for ReDial and JDDC.
Following~\citep{Peng2020SOLOISTFT}, we generate five delexicalized candidates~\citep{Nekvinda2021ShadesOB} through nucleus sampling, choose the one that contains more slots in the utterance, and back-fill the slots with the retrieval results from the recommender. 
\changed{The code of tester-based framework is available at \url{https://github.com/Superbooming/simtester}.}

\section{Evaluation and Results}
\label{sec:res}
\sww{
In this section, we describe how we evaluate the simulators and testers.
\begin{itemize}
    \item \textbf{Evaluating simulators:} To evaluate the realism of the user simulators, we examine the likeness of simulator-generated responses to human responses, and evaluation metrics including word-overlap rate, diversity, entity accuracy, conversation success rate, and human evaluation are employed; see Sect. \ref{sec:sec:eval-sim} for details.
    \item \textbf{Tester-based evaluation:} To evaluate the capability of simulators to distinguish between systems, we rely on the newly proposed tester-based evaluation framework to test the ability of the simulator to discriminate between dialogue systems with different capabilities; see Sect. \ref{sec:sec:eval-testers} for details.
    \item \textbf{Evaluating testers:} To validate the reliability of the testers of the tester-based evaluation, we measure the performance of the dialogue system under the tester definition in dialogue benchmarks, interactive tests, and human evaluation; see Sect. \ref{sec:sec:reliability} details.
\end{itemize}
In addition to this, we conduct a series of analytical experiments, including an in-depth comparison on agenda-based and model-based simulators, efficiency evaluation, human evaluation, and support for evaluating third-party dialogue systems; see Sect. \ref{sec:sec:analytical} for details.
}


\subsection{\sww{Evaluating simulators}}
\label{sec:sec:eval-sim}

\header{Evaluation methods}
\sww{We evaluate the simulators by evaluating the response generated by the simulators and the quality of interactions following three strategies:}
\begin{itemize}
    \item \textbf{Test set evaluation} evaluates the simulators on the test set. Using the following metrics, we measure the next user utterance generated based on dialogue context by user simulators:
    \textsc{F1}\footnote{\url{https://github.com/facebookresearch/ParlAI/blob/master/parlai/core/metrics.py}} is a text generation metric that measures the word overlap between the generated responses and ground-truth responses. 
    \textsc{Distinct}~\citep{Li2016ADO} estimates the proportion of distinct 3-gram in the data, which indicates the lexical diversity.
    \textsc{SlotAcc} measures the slot accuracy. \textsc{SlotAcc} equals $1$ if all the slots (or entities) in the ground-truth response appear in the generated text, otherwise $0$.
    
    \item \textbf{Interaction evaluation} measures the interaction quality of the simulators. The simulators generate 10K dialogues by interacting with the \emph{base system}, and are evaluated by the \textsc{Success} rate and \textsc{Diversity} of the simulated dialogues.
    \textsc{Success}~\citep{Eric2020MultiWOZ2A} measures if the recommended item meets all user preferences (we discard \textsc{Success} for JDDC due to the lack of definition).
    \textsc{Distinct} measures the diversity of simulated dialogues.
    
    \item \textbf{Human evaluation} evaluates the simulators by the human assessment on simulated dialogues. 
    We ask ten workers to rate each of the simulated dialogues in terms of the \textsc{Coherence} and \textsc{Human-like}~\citep{Shi2019HowTB}. 
    Specifically, we randomly sample about 700 dialogues from the simulated data and have people rate a \textsc{Coherence} in range 1 to 3, as well as a \textsc{Human-like} score in range 1 to 3 (higher scores are better.). 
    
\end{itemize}

\header{Results}
The experimental results are provided in Table~\ref{table:sim-test} and Table~\ref{table:sim-human}.
MetaSim consistently outperforms the baseline methods in terms of \textsc{SlotAcc}, which indicates that MetaSim can inform the system about individual requirements more accurately.
MetaSim exceeds Seq2seq on \textsc{SlotAcc} with an absolute improvement of +8.24 in MultiWOZ, +4.47 in ReDial and +6.47 in JDDC. 
The rule-based methods (e.g., Agenda) perform worse than data-driven methods on the test set, especially on JDDC, which indicates that human-curated rules are difficult to cover all the conversation situations.
MetaSim also outperforms baselines in terms of \textsc{F1}: the responses generated by MetaSim are more natural than baseline methods.
In terms of \textsc{Distinct}, MetaSim surpasses all generative methods but it is inferior to the retrieval-based techniques (i.e., Agenda and SL+Template) in MultiWOZ and ReDial. 
The retrieved responses are more diverse since they are based on templates extracted from human-curated responses from the corpus.
Note these responses may not be appropriate for the current conversation. 

The experiments on Interaction evaluation indicate that
MetaSim achieves a higher task success rate than the baselines, demonstrating better conversational skills (understanding system questions and informing system requirements). 
MetaSim also exceeds the baselines in terms of diversity except for the agenda model on ReDial data. 
The human evaluation shows that the simulated dialogue by MetaSim is more coherent to humans and human-like than the baselines.\footnote{We do not obtain a significant improvement in the human evaluation due to the limited number of annotations. We consider a large-scale human evaluation as future work.}
The ablation experiments show that removing modules brings a consistent decrease in performance. 
Specifically, MetaSim beats MetaSim-Metaphor, indicating a better dialogue ability and diversity via adding the module of Metaphor.
Similarly, removing the Policy module leads to a clear performance drop. 
The MetaSim-Preference is less capable of completing a task-oriented dialogue without preference constraints and a clear task goal.

\begin{table}[!t]
\centering
\caption{Results of tester-based evaluation. $\mathbb{T}_C$ denotes the Context tester, $\mathbb{T}_R$ denotes the Recommender tester, and $\mathbb{T}_D$ denotes the Domain tester. Human evaluation results are only included for MultiWOZ as detailed in Sect.~\ref{sec:sec:eval-testers}.
}
\label{table:tester}
\begin{tabular}{l ccc ccc cc}

\toprule

{\textbf{Datasets}} 
& \multicolumn{3}{c}{\textbf{MultiWOZ}} 
& \multicolumn{3}{c}{\textbf{ReDial}} 
& \multicolumn{2}{c}{\textbf{JDDC}}\\

\cmidrule(lr){2-4} \cmidrule(lr){5-7}  \cmidrule(lr){8-9}  

\textbf{Tester} & $\mathbb{T}_C$     & $\mathbb{T}_R$   &  $\mathbb{T}_D$
& $\mathbb{T}_C$     & $\mathbb{T}_R$   &  $\mathbb{T}_D$
& $\mathbb{T}_C$     &  $\mathbb{T}_D$
\\ 

\midrule

Agenda
& 16.93     & 17.28   &  19.47
& 16.41     & 17.12   &  17.05
& 13.76     & 14.52
\\

Seq2seq
& 30.84     & 37.15   &  32.81
& 16.52     & 17.46   &  19.16
& 18.47     &  16.89
\\

\textbf{MetaSim}
& \textbf{33.23}     & \textbf{40.11}   &  \textbf{35.21}
& \textbf{17.48}     & \textbf{20.61}   &  \textbf{20.43}
& \textbf{19.17}     & \textbf{17.47}
\\
\midrule

Human
& 43.63    & 40.54  &  42.54
& --     & --   &  --
& --    & -- 
\\

\bottomrule
\end{tabular}
\end{table}

\subsection{\sww{Tester-based evaluation}}
\label{sec:sec:eval-testers}

\header{Evaluation methods}
We conduct experiments with our testers and compare several interactive evaluators (i.e., simulators and humans) in terms of \textsc{ExactDistinct} to evaluate the simulators.
Following the strategies described in Sect. \ref{sec:sec:tester}, we implement three testers for each of the three datasets except for JDDC (cf. Sect.~\ref{sec:sec:iml-tester}).
We initialize the preferences from the preference module and let the simulators interact with system variants defined by  testers.
The simulators give a rating of the system after the interaction.
We use user calibrated user satisfaction, 
i.e., the average of satisfaction and task success (as described in Sect.~\ref{sec:sec:inference}) as the ratings of models.
If the scoring of the two system variants is the same, we further distinguish them by the number of dialogue turns (the one with fewer rounds is better).
We then rank the variants based on the ratings and check if the order is the same as expected (i.e., $S_1<S_2<S_3$). The simulator will be scored 1 on \textsc{ExactDistinct} if the order given by the simulator is consistent with expectations, otherwise scored as $0$.
We repeated the testing on 1K different preferences and calculated the averaged \textsc{ExactDistinct} score.

We also conduct a human evaluation on MultiWOZ to validate the reliability, i.e., if the human can distinguish the system variants. 
Specifically, we develop an annotation system and ask 40 annotators to talk with the system variants to complete a specific goal. 
We randomly assign systems and goals to users to avoid bias caused by order of operations.
After the dialogue, the annotator is asked to rate the system in terms of  \textsc{Satisfaction} and \textsc{Success}. 
We use the averaged \textsc{Satisfaction} and \textsc{Success} as the ratings of the system.

\header{Results}
Table~\ref{table:tester} shows the results of the tester evaluation. 
First, MetaSim is more capable of distinguishing the system variants defined in a tester and achieving the best in terms of the \textsc{ExactDistinct} score than Agenda model and Seq2seq model.
This might be because the agenda model is limited to domain-specific human-curated rules and difficult to generalize, and end-to-end simulators can only evaluate the systems on the semantic level.
Second, the difficulty of the tester varies, as does the difficulty of the tester on different data (e.g., Recommender tester is more difficult for people than others, because the responses generated by system defined by Recommender tester look reasonable and may confuse users). 
Third, people can better distinguish between different systems compared to simulators (e.g., Human get an \textsc{ExactDistinct} of 43.63, 40.54, and 42.54 for the three testers on MultiWOZ, compared to 33.23, 40.11, and 35.21 for MetaSim).
Note there still exists a performance gap between MetaSim and Humans, indicating that building human-like simulators is still a challenge. We leave it for future work of investigating the reasons behind it.

\begin{table}[!t]
\centering
\caption{Evaluation results of variants on the MultiWOZ dataset. Dist denotes \textsc{Distinct}, Sat. denotes \textsc{Satisfaction}, Eff. denotes \textsc{Efficiency}, and Nat. denotes \textsc{Naturalness}.}
\label{table:system-mwoz}
{%
\begin{tabular}{@{} l  ccc cc cccc @{}}
\toprule
{}
& \multicolumn{3}{c}{Test Set} 
& \multicolumn{2}{c}{Interaction}
& \multicolumn{4}{c}{Human}
\\
\cmidrule(r){2-4}
\cmidrule(r){5-6}
\cmidrule{7-10}
& BLEU     & Success   &  Slot  & Success  & Dist   & Sat.   & Success & Eff. & Nat.

\\ 

\midrule

System
& 20.42 &  75.00  &51.92  &  80.25   & 15.32    
&  3.54   &   39.50 &  84.57 & 2.37

\\


\midrule

\ $\alpha=3$
&18.84  & 57.65  &51.75  & 40.20     & 13.76   
& 3.07    &   29.09 &  74.54 & 2.15

\\

\ $\alpha=1$
&16.81  &46.70  &44.59  &  27.04  & 15.16    
& 2.32    &   12.20 &   51.22 & 1.66
 
\\


\midrule

\ $\beta=0.4$
& 20.49    & 38.17    & 51.69    & 28.60    &  15.27   
& 3.26    &  24.53  &  77.36 & 2.13
  
\\

\ $\beta=0.1$
& 20.53    & 34.90    & 52.33    & \phantom{8}7.40    & 10.17    
&  3.24   &   13.33 &  71.11 & 2.07

\\

\midrule
\ $\gamma=0.1$
&  16.10   & 42.53    & 33.66    & 31.70    & 15.84    
&  3.10   &   16.33  & 75.51 & 2.06

\\

\ $\gamma=0.01$
& 13.18    & 38.54    & 20.02    & 25.30    & 14.51    
&  2.55   &  \phantom{8}2.13  & 59.57 & 1.89
\\
\bottomrule
\end{tabular}
}
\end{table}

\begin{table}[!t]
\centering
\caption{Evaluation results of variants on the ReDial and JDDC datasets; same conventions as in Table~\ref{table:system-mwoz}. No results are included for the $\beta$ (recommender tester) for JDDC as JDDC does not have a recommender component due to the unavailability of its database (cf.~Sect.~\ref{sec:sec:iml-tester}).}
\label{table:system-redial}{%
\begin{tabular}{@{}l  ccccc cccc @{}}
\toprule

{\textbf{Datasets}} 
& \multicolumn{5}{c}{\textbf{ReDial}} 
& \multicolumn{4}{c}{\textbf{JDDC}}\\

\cmidrule(lr){2-6} \cmidrule(lr){7-10}  

{}
& \multicolumn{3}{c}{Test Set} 
& \multicolumn{2}{c}{Interaction}
& \multicolumn{3}{c}{Test Set} 
& \multicolumn{1}{c}{Interaction} 
\\
\cmidrule(r){2-4}
\cmidrule(r){5-6}
\cmidrule(r){7-9}
\cmidrule{10-10}

& BLEU     & Success   &  Slot  & Success  & Dist   
& BLEU     & Dist   &  Slot &  Dist  
\\ 

\midrule

System
& 15.42  & 8.20    & 3.74    &  13.70   & 20.97
& 25.45    & 10.68    & 8.72    &    34.81

\\


\midrule

\ $\alpha=3$
&14.87  & 8.20    & 3.37    &  11.45   &   23.01  
& 21.76    &  \phantom{1}9.49    & 6.80       & 31.10

\\

\ $\alpha=1$
&14.34  & 6.93    &  2.91   &  \phantom{1}9.62   & 21.13
& 14.64    & \phantom{1}6.84    & 1.14      &  29.74  
\\


\midrule

\ $\beta=0.4$
& 15.44     & 1.86     & 2.96     &  \phantom{1}6.54   &  21.43   
&  --   &  --   &  --   & --       
\\

\ $\beta=0.1$
& 14.80     & 0.75     & 2.75     &  \phantom{1}2.45   & 27.14
& --    &  --   &  --   &  --
\\

\midrule
\ $\gamma=0.1$
&  14.91   & 4.62    & 3.03    &  \phantom{1}5.95   &   18.52   
& 22.48    & 10.05    & 6.18    &  33.68 
\\

\ $\gamma=0.01$
& 13.99     & 1.19    & 1.45    & \phantom{1}2.27   &   17.26
& 17.5\phantom{0}    & \phantom{1}9.94    & 1.92    &  34.25  
\\
\bottomrule
\end{tabular}
}
\end{table}

\subsection{\sww{Evaluating testers}}
\label{sec:sec:reliability}

\textbf{Evaluation methods.}
\sww{To assess the variants of these testers with sensible performance differences, we evaluate their performance in the test set, interaction quality, and human assessment.}
\begin{itemize}
    \item \textbf{Test set evaluation} evaluates the system variants defined by testers on the test set. Specifically,
    similar to the way to evaluate simulators, we measure the generated utterances using the following metrics:
    \textsc{BLEU}~\citep{Papineni2002BleuAM} measures the word-overlap between the generated utterance the ground-truth utterance.
    \textsc{Success} measures if the recommended item meets all user preferences.
    \textsc{SlotAcc} measures the slot accuracy.
    
    \item \textbf{Interaction evaluation} evaluates the system by interacting with MetaSim. We generate 10K dialogue and calculate the \textsc{Satisfaction} and \textsc{Distinct}. 
    \textsc{Success} measures if the system recommend an item that meets all the user needs.
    \textsc{Distinct} measures the diversity of simulated dialogue. 
    
    \item \textbf{Human evaluation} evaluates the systems by human experiments. Human evaluation was operated on the annotation system we built for MultiWOZ and invited 40 users to talk to the system based on the predefined preference. 
    We collected about 500 dialogues, where the users were instructed to label the following aspects at the end of each dialogue:
    \begin{enumerate*}[label=(\roman*)]
    \item \textsc{Success} measures if the system solves the user's problem, rating as: $1={}$Fail, $2={}$Success;
    \item \textsc{Efficiency} measures the efficiency of the system, rating as: $1={}$Inefficient, $2={}$Efficient;
    \item \textsc{Naturalness} measures the human-likeness of the system, rating as: $1={}$Unnatural, $2={}$Fair, $3={}$Natural;
    \item \textsc{Satisfaction} is labeled on a 5-point scale: $1={}$Very bad, $2={}$Bad, $3={}$Fair, $4={}$Good, $5={}$Very good.
    \end{enumerate*}
\end{itemize}

\header{Results}
Table~\ref{table:system-mwoz} shows the evaluation results on MultiWOZ and Table~\ref{table:system-redial} shows the evaluation results on ReDial and JDDC.
\changed{We have three main findings from the results.}

First, the differences between system variants can be distinguished by humans. 
The user satisfaction annotations (as an overall rating for the systems) are consistent with our expectations, while other human evaluation metrics demonstrate the same results but with a focus on the specific capabilities of the system. 
\changed{
We also notice a smaller performance gap between the recommender tester-defined human evaluation variants and the other two testers.
This is because of human annotators' difficulties judging the goodness of the plausible-looking results recommended by different systems, a finding consistent with that in tester evaluation;  see Sect.~\ref{sec:sec:eval-testers}.
}

Second, each tester has focused on a specific capability of the system. For example, the recommender tester largely disrupts the system's ability to make recommendations (low \textsc{Success} in both test set, interaction, and human evaluation), while keeping the ability to generate natural responses (relatively good \textsc{BLEU} and \textsc{Naturalness}).
\changed{
Therefore, we measure a user simulator's capabilities to evaluate the dialogue system's recommendation capabilities by checking whether it can accurately distinguish the differences between the system variants defined by the recommender tester.
}

Third, the interaction experiments also exhibit results consistent with expectations, but the difference between good and bad systems are more significant. 
This could be attributed to the accumulation of errors in the interaction amplifies the difference in system performance.
\changed{
For example, in Table~\ref{table:system-mwoz}, we see that the \textsc{Success} gap between the base system and system variant with $\alpha=3$ in interaction evaluation is more significant than that of in test set evaluation ($75.00-46.70=28.30$ compared to $80.25-27.04=53.21$).
We notice that the diversity (\textsc{Distinct}) of the variants do not follow the rule that ``\emph{weaker variants perform worse}''.
For example, in Table~\ref{table:system-mwoz} we see that the system variant with $\alpha=1$ achieves a higher \textsc{Distinct} score ($15.16$) compared to the system variant with $\alpha=3$ ($13.76$)).
This is possibly because the designed testers lacked control over response generation diversity.
}



\subsection{Analytical experiments}
\label{sec:sec:analytical}
\changed{
We conduct further analytical experiments to understand the results better.
}

\begin{table*}[!t]
\centering
\caption{\changed{Result comparison between agenda-based and model-based user simulators on MultiWOZ. The top block shows the results of different variants of agenda-based user simulators. 
The bottom block shows model-based user simulators (MetaSim). The numbers in parentheses represent the number of training dialogues.}\sww{* indicates statistically significant improvements over baselines (p-value $< 0.05$).}}
\label{table:good-agenda}
{%
\begin{tabular}{l ccc cc cc}

\toprule

{\textbf{Datasets}} 
& \multicolumn{3}{c}{\textbf{Test set}} 
& \multicolumn{2}{c}{\textbf{Interaction}} 
& \multicolumn{2}{c}{\textbf{Human}}\\

\cmidrule(lr){2-4} \cmidrule(lr){5-6}  \cmidrule(lr){7-8}  

& F1     & Dist.   &  Slot  
& Success  & Dist.  
& Coh.     & Hum. 
\\ 

\midrule
\emph{Agenda-based models}\\


ConvLab2+Template~\citep{Zhu2020ConvLab2AO}
& 18.26 & \phantom{0}5.77 & 18.25 & 27.10 & \phantom{0}7.28 & 1.65 & 1.74
\\

ConvLab2+Retrieval~\citep{Zhu2020ConvLab2AO}
& 18.39 & \textbf{18.66} & 18.70 & 18.10 & \textbf{20.47} & 1.46 & 1.84
\\

ConvLab2+Template (oracle)~\citep{Zhu2020ConvLab2AO}
& 20.00 & \phantom{0}5.91 &  21.39 & - & -  & - & -
\\

ConvLab2+Retrieval (oracle)~\citep{Zhu2020ConvLab2AO}
& 19.73 & 18.53 & 21.42 & - & - & - & -
\\


\midrule
\emph{Model-based models}\\
MetaSim
& \textbf{44.13}\rlap{$^{*}$} & 17.29 & \textbf{57.65}\rlap{$^{*}$} & \textbf{80.25}\rlap{$^{*}$} & 15.65 & \textbf{2.51} & \textbf{2.61}
\\
MetaSim (1k)
& 41.68\rlap{$^{*}$} & 14.75 & 50.56\rlap{$^{*}$} & 59.30\rlap{$^{*}$} & 16.31 & 2.23 & 2.51
\\
MetaSim (100)
& 35.86\rlap{$^{*}$} & 12.30 & 31.60\rlap{$^{*}$} & 35.90\rlap{$^{*}$} & 13.56 & 1.80 & 2.10
\\
\bottomrule
\end{tabular}
}

\end{table*}

\header{Comparison of agenda-based and model-based user simulators}
User simulators fall into two main categories, agenda-based user simulators and model-based user simulators~\citep{Shi2019HowTB,Zhu2020ConvLab2AO}. 
To bootstrap the agenda-based user simulator, we might use human-defined rules, such as a manually designed state transition matrix, for updating dialogue policy.
In contrast, model-based user simulators learn dialogue policy from the corpus.
In Sect~\ref{sec:sec:eval-sim}, we define the state transition matrix of the agenda-based simulators by statistically measuring the distribution of the training dataset.
Here, we compare MetaSim with the agenda-based user simulators implemented in ConvLab2~\citep{Zhu2020ConvLab2AO}, \sww{which are state-of-the-art multi-domain agenda-based simulators for MWOZ and have been included as the standard test kits for DSTC competitions\footnote{\url{https://github.com/thu-coai/ConvLab-2}}.}
The agenda-based user simulators implemented in ConvLab2 consist of three modules: NLU, policy, and NLG. 
The NLU module employs a BERT-based model trained on MultiWOZ to identify system actions. 
The policy module employs sophisticated, manually defined rules to produce user actions.
The NLG module converts user actions into responses; two strategies are employed: \emph{Template} handwrite templates for each action, and \emph{Retrieval} use actions as keywords to retrieve relevant responses in MultiWOZ training data and replace entities.
We denote the simulators as \emph{ConvLab2+Template} and \emph{ConvLab2+Retrieval}, depending on their NLG strategies.
We encounter \emph{order inconsistencies} between the agenda built automatically by agenda-based simulators and the agenda of the real dialogue in test set evaluation. 
For example, assume a user plans to complete two tasks in conversation, booking a restaurant and booking a train by order.
However, the simulator might automatically build the agenda to book the train first and then book the restaurant.
It may be able to accomplish two tasks as well.
The order inconsistency of agenda order would degrade the performance of agenda-based simulators in the test set evaluation. 
Therefore, we also evaluate the agenda-based simulator under \emph{oracle condition}, i.e., we initialize the simulator's agenda with the agenda of the real conversation in the test set evaluation.
We did not employ the agenda-based simulators designed by \citet{Shi2019HowTB} because they only developed rules for the restaurant reservation domain.

\begin{figure}[t]
 \centering
 \includegraphics[width=0.7\columnwidth]{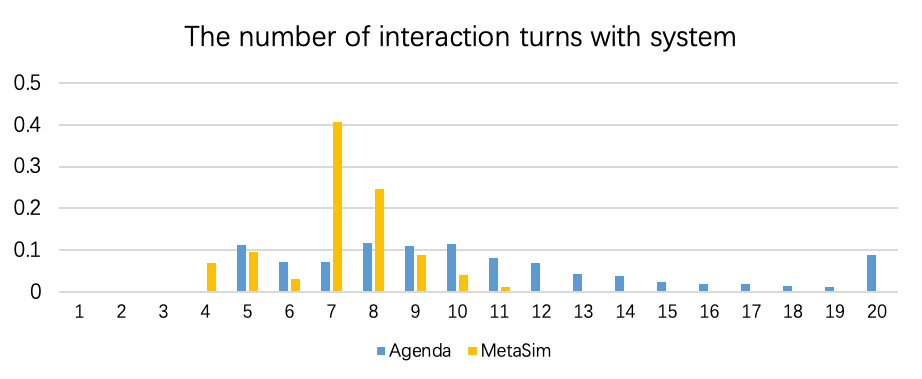}
 \caption{The number of turn of interaction between base system and user simulators.
 }
 \label{fig:turn}
\end{figure}

We evaluate the performance of ConvLab2 simulators and MetaSim regarding test set, interaction quality, and human assessment.
We also include two variants of MetaSim in evaluation: MetaSim (1k) and MetaSim (100), which are MetaSim models trained using 1k annotated dialogues and 100 annotated dialogues, respectively.
The results are listed in Table~\ref{table:good-agenda}. 
Note that human evaluation in Table~\ref{table:sim-human} and Table~\ref{table:good-agenda} are two independent experiments in which the participating annotators, the evaluated dialogues, and the instruction are different; therefore, the absolute values of their results are different.
From the results, we have two main findings:
(i) The model-based user simulators significantly outperform the rule-based (e.g., agenda-based) simulators. We find that the model-based user simulators can generate more contextual responses and have a higher success rate when interacting with the system. Their interactive dialogues are considered more human-like in human evaluations.
\sww{
To further compare the behavior of agenda-based methods and our proposed method, we listed the number of interaction turns between the base system and user simulators in Figure~\ref{fig:turn}.
From the results, MetaSim's interaction rounds are within ten rounds. In comparison, the ConvLab2 Agenda model's interaction rounds have more than 20 rounds, showing the advantage of MetaSim in efficiency.
}
(ii) The model-based simulator performs better when using a small amount of annotated data. For example, MetaSim (100) outperforms agenda-based simulators using 100 annotated dialogues.
(iii) Retrieval-based models, such as ConvLab2+Retrieval, outperform generation-based models in terms of diversity.
\sww{
This could be because retrieval-based methods pick human-written responses under the predicted action more randomly from the training. In contrast, generative methods generate similar responses when presented with the same action.
Additionally, using a sampling-based decoding method (like nucleus sampling) can enhance the diversity of the generative model. 
We plan further to enhance the model generation's diversity in future work.
}

\begin{table*}[!t]
\centering
\caption{\changed{Evaluation results of MetaSim variants on MultiWOZ.} \sww{* indicates statistically significant improvements over Seq2seq method (p-value $< 0.05$).}}
\label{table:efficiency}
{%
\begin{tabular}{l cc ccc cc}

\toprule

& Storage   & Time   &  F1  & Distinct  & Slot & Success & Interact-Distinct
\\ 

\midrule
Seq2seq & 0.89 & \phantom{0,}409 & 42.02 & 14.98 & 49.41 & 76.75 & 13.76
\\
MetaSim & 3.56 & 1,987 & 44.13\rlap{$^{*}$} & \textbf{17.29}\rlap{$^{*}$} & 57.65\rlap{$^{*}$} & 80.25 & \textbf{15.65}
\\
\midrule
Variant A
& 2.67 & 1,250  & \textbf{44.56}\rlap{$^{*}$} & 16.64\rlap{$^{*}$} & \textbf{58.32}\rlap{$^{*}$} & \textbf{81.10} & 14.99
\\
Variant B
& 0.89 & \phantom{0,}657 & 43.15\rlap{$^{*}$} & 14.76 & 54.46\rlap{$^{*}$} & 77.40 & 10.97
\\
Variant C
& 0.92 & \phantom{0,}471 & 43.99\rlap{$^{*}$} & 10.57 & 57.39\rlap{$^{*}$} & 79.60 & 11.88
\\
\bottomrule
\end{tabular}
}

\end{table*}

\header{The efficiency of MetaSim} The proposed MetaSim uses modular modeling, i.e., different parts (NLU, Policy, etc.) are modeled separately using different T5 models. The ablation study in Table~\ref{table:sim-test} shows the model's performance with some of the modules removed. Here, we investigate whether specific modules can be consolidated together to improve efficiency and how these changes affect the model's performance.
The tested model variants are:
(i) \emph{Variant A: Consolidate policy and NLG modules.} We use one T5 model to sequentially predict the action $a_{t+1}$ and the response $r_{t+1}$. The action and response are concatenated with a special token \code{</s>}.
(ii) \emph{Variant B: Consolidate NLU, policy, and NLG modules in one T5 model.} This variant consolidates the NLU, policy, and NLG modules in one T5 model by sequentially generating system action, user action, and user response. The prediction of the metaphor module is removed.
(iii) \emph{Variant C: MetaSim with T5-small.} This variant replaces the T5-base model with T5-small.

The model's results on MultiWOZ variants are listed in Table~\ref{table:efficiency}. The metrics are the Storage space of the model (GB), inference time (seconds) on the MultiWOZ test set, F1, Distinct, and SlotAcc of the generated response.
We can see that variants A and B reduce the model storage space and computational overhead by merging the modules and achieving very competitive results.
\sww{For example, variant A performs comparably to MetaSim by mildly outperforming in terms of F1 and Slot and lags behind MetaSim in Distinct; this may be because the action prediction task and the response generation task share similar capabilities and can be done using a single model.}
Variant C, which replaces the T5-base with T5-small, also significantly reduces computational overhead: it is $4\times$ faster than MetaSim with the T5-base model.
Variant C also achieves competitive results in terms of F1 and Slot but is ineffective in generating diverse responses (e.g., low Distinct).
\sww{This may be due to model capacity limitations whereby variant C generates generic responses.}


\begin{table*}[!t]
\centering
\caption{\changed{Evaluation results of human evaluation. We compare three settings: 3-point scale scoring, 5-point scale scoring, and rank-based evaluation. 
Score denotes the average scores of the models. 
Best and Worst measure the number of cases in which the corresponding models perform the best or worst among all the models.}}
\label{table:scale}
{%
\begin{tabular}{l ccc ccc ccc}

\toprule
{\textbf{Datasets}} 
& \multicolumn{3}{c}{\textbf{3-Scale Score}} 
& \multicolumn{3}{c}{\textbf{5-Scale Score}} 
& \multicolumn{3}{c}{\textbf{Rank}}
\\

\cmidrule(lr){2-4} \cmidrule(lr){5-7}  \cmidrule(lr){8-10}  

&  Score & Best  & Worst 
&  Score & Best  & Worst 
&  Score & Best  & Worst 
\\ 

\midrule

Agenda
& 1.33    &  \phantom{0}27   & 255 
& 1.93  & \phantom{0}15 & 248
& 1.13 & \phantom{00}5 & 239
\\

Seq2seq
& 2.31    &  160   & \phantom{0}84 
& 3.53 & 105  & \phantom{0}52 
& 2.20 & \phantom{0}78 & \phantom{0}25
\\

MetaSim
& 2.72 & 244   & \phantom{0}32
& 4.15 & 210  & \phantom{0}16
& 2.67 &  187 &  \phantom{00}7
\\
\midrule
Kappa
& \multicolumn{3}{c}{34.56}
& \multicolumn{3}{c}{22.78}
& \multicolumn{3}{c}{60.00}
\\
\bottomrule
\end{tabular}
}

\end{table*}

\header{Human evaluation methods}
In previous experiments, we use 3-scale in human evaluation in Table~\ref{table:sim-human}. Here we compare the different human evaluation methods, they are:
(i)\emph{Score-based evaluation in 3-scale.} 
Human annotators are asked to rate a score to the responses in range 1-3.
(ii) \emph{Score-based evaluation in 5-scale.} 
Human annotators are asked to rate a score to the responses in range 1-5.
(iii) \emph{Rank-based evaluation.}
Human annotators are asked to rank multiple responses of different models with respect to the dialogue context.
The metrics we use are: \emph{Score}, the average score of the model; \emph{Best}, the number of examples that the model performs the best (or tied for the best) among models; \emph{Worst}, the number of examples that the model performs the worst (or tied for the worst) among models.
We also calculate the \emph{Kappa}, i.e., the agreement of the three human annotators of the three evaluation methods.
Table~\ref{table:scale} lists the results.
MetaSim is considered more realistic than baseline models Agenda and Seq2seq across different human evaluation methods: it obtains higher average scores and is considered the best model in more examples.
We also find that the agreement of 3-scale scoring is better compared to 5-scale scoring, while the agreement of the rank-based evaluation method is the best.

\begin{table*}[!t]
\centering
\setlength\tabcolsep{1pt}
\caption{\changed{Evaluation results of dialogue systems on MultiWOZ.}
}
\label{table:3-party}
{%
\begin{tabular}{l  cccc cccc ccc}

\toprule
{}
& \multicolumn{4}{c}{Benchmark}
& \multicolumn{4}{c}{Interaction}
& \multicolumn{3}{c}{Human}
\\

\cmidrule(lr){2-5} \cmidrule(lr){6-9} \cmidrule(lr){10-12}
& Inform & Success & BLEU & Combine
& Success & Distinct & Sat. & Combine
& Sat. & Effic. & Natural.
\\ 

\midrule


DAMD~\citep{Lei2018SequicityST}
& 76.40 & 60.40 & 16.60 & 85.00 & 20.90 & 10.14 & 2.15 & 0.392  & 1.79 & 1.88 & 1.87
\\

SOLOIST~\citep{Peng2020SOLOISTFT}
& 85.50 & 72.90 & 16.54 & 95.74 & 71.00 & \textbf{14.23} & 2.74 & 0.790 & 2.40 & 2.54 & 2.52
\\

GALAXY~\citep{He2021GALAXYAG}
& \textbf{95.40} & \textbf{80.70} & \textbf{17.00} & \textbf{105.10} & \textbf{74.90} & 9.59 & \textbf{2.78} & \textbf{0.819}  & \textbf{2.66} & \textbf{2.58} & \textbf{2.57}
\\

\bottomrule
\end{tabular}
}

\end{table*}

\header{Test results of third-party dialogue systems}
We used SOLOIST as the base dialogue system in the experiments described above.
Here, we conduct experiments on some third-party dialogue systems, which are:
(i) \emph{DAMD}~\citep{Zhang2020TaskOrientedDS}, a modular \ac{TDS} system where each dialog module is implemented using a neural network, and the whole system is trained in an end-to-end manner;
(ii) \emph{SOLOIST}~\citep{Peng2020SOLOISTFT}, a \ac{TDS} models pre-trained on heterogeneous dialog corpora;
(iii) \emph{GALAXY}~\citep{He2021GALAXYAG} a pre-trained \ac{TDS} model that explicitly learns dialog policy from limited labeled dialogues and large-scale unlabeled dialog corpora via semi-supervised learning.
The evaluation results are listed in Table~\ref{table:3-party}. 
We can see that:
(i) GALAXY performs the best in three evaluation methods, while DAMD performs the worst.
(ii) User simulators (i.e., MetaSim) have the ability to discriminate between different capability systems, and their predicted model success rate, user satisfaction, and combined scores (i.e., an average of success and satisfaction), all meet expectations.
\sww{
Table~\ref{table:3example} illustrates an example of a conversation between MetaSim and three dialogue systems, in which we can see that DAMD did not understand the user's needs when they first asked for a postcode.
SOLOIST and GALAXY performed well, but GALAXY had some expression repetition problems.
}

\begin{table*}[!t]
\centering
\setlength\tabcolsep{4pt}
\caption{Example of dialogues using DAMD, SOLOIST and GALAXY.} \label{table:3example}
\resizebox{\columnwidth}{!}{
\begin{tabular}{@{}rp{17cm}@{}}
\toprule

\textbf{Goal} & \underline{You are planning your trip in Cambridge. You are looking for a place to stay. The hotel should be in the cheap price } \underline{range and should include free parking. The hotel should include free wifi.} You are also looking for a train. The train should arrive by 09:30 and should go to cambridge. The train should leave on monday and should depart from london kings cross.\\

\midrule
\textbf{Model} &{DAMD}\\
\midrule
user:& i'm looking for a cheap place to stay that offers free parking.\\
system:&  there are 22 restaurants that fit your criteria . do you have a preference for area ?\\
user:&   no, but it should also have free wifi.\\
system:& i would recommend the pizza hut city centre . it is in the centre .\\
user:&   could i get the postcode?\\
system:& pizza hut city centre is a italian restaurant in the centre . it is in the cheap price range .\\
user:&   i think that is good. can i get the post code, please?\\
system:& the postcode is cb21ab . their phone number is 01223323737 . is there anything else i can help you with ?\\
user: & ...\\

\midrule
\textbf{Model} &{SOLOIST}\\
\midrule
user:&   i'm looking for a place to stay that includes free wifi and free parking.\\
system:& there are 29 places that meet that criteria . is there a particular area you would like to stay in ?\\
user:&   no particular area, but i would like it to be cheap.\\
system:& i have 10 guesthouses that fit that criteria . 5 in the centre and 2 in the east . do you have a preference ?\\
user:&   i would like the one in the east please.\\
system:& allenbell is a cheap guesthouse in the east with free wifi and parking . would you like to book a room ?\\
user:&   no, but can you give me the postcode?\\
system:& the postcode for allenbell is cb13js . can i help you with anything else today ?\\
user: & ...\\

\midrule
\textbf{Model} &{GALAXY}\\
\midrule
user:&   i'm looking for a cheap place to stay that offers free parking.\\
system:& i have 10 places that fit that criteria . is there a particular area you would like to stay in ?\\
user:&   no, but it should also have free wifi.\\
system:& i have 10 places that meet that criteria . is there a particular area you would like to stay in ?\\
user:&   i don't have a particular area. which would you recommend?\\
system:& i would recommend the alexander bed and breakfast in the centre . would you like me to book a room for you ?\\
user:&   i don't need to book a room, but could you tell me the hotel type, phone number, and address please?\\
system:& it is a guesthouse located at 56 saint barnabas road . the phone number is 01223525725 .\\
user: & ...\\
\bottomrule
\end{tabular}
}
\end{table*}

\begin{table*}[!t]
\centering
\caption{Evaluation results on unseen domain. \textbf{Boldface} indicates significant improvements over baselines with (p-value $< 0.05$)}
\label{table:unseen}
{%
\begin{tabular}{l ccc ccc }

\toprule

{\textbf{Datasets}} 
& \multicolumn{3}{c}{Hotel} 
& \multicolumn{3}{c}{Attraction} \\

\cmidrule(lr){2-4} \cmidrule(lr){5-7}

& F1     & Dist.   &  Slot  
& F1     & Dist.   &  Slot  
\\ 

\midrule

Seq2seq
& 34.21 & 8.16 & 33.11
& 37.26 & 8.50 & 39.05

\\

MetaSim
& \textbf{39.73}    &  \textbf{12.41}   & \textbf{39.79}
& \textbf{41.45}    & \textbf{12.69}    & \textbf{50.12}
\\

\midrule

MetaSim-Metaphor
& 35.63    &  10.70   & 37.38
& 38.07    &  10.58   & 47.66
\\

MetaSim-Ranking
& 37.89    &  11.80   & 37.00
& 40.02   &  12.36   & 47.72
\\

\bottomrule
\end{tabular}
}

\end{table*}

\header{Test results on unseen domains}
\sww{
We further conducted experiments on unseen domains. We removed all samples, including ``hotel'' or ``attraction'' domains in the training set, respectively, trained the model, and tested it on the subset of the test set containing ``hotel'' and ``attraction''.
Table~\ref{table:unseen} shows the results.
We observe that the proposed model achieves the best results.
MetaSim-Metaphor, a MetaSim variant that does not use the retrieved records, shows a clear drop in effectiveness. MetaSim-Ranking, the variant with the ranking module removed from the metaphor module, had a noticeable drop compared to MetaSim, especially in the Slot accuracy metric.
Finally, compared to the Seq2seq model, MetaSim shows a significant improvement, possibly due to its ability to respond more appropriately to items of new domains by retrieving known similar records. For instance, many behaviors in hotel reservations are similar to known restaurant reservation records, and a reasonable association can improve the model's response.
}

\begin{figure}[t]
 \centering
 \includegraphics[width=0.9\columnwidth]{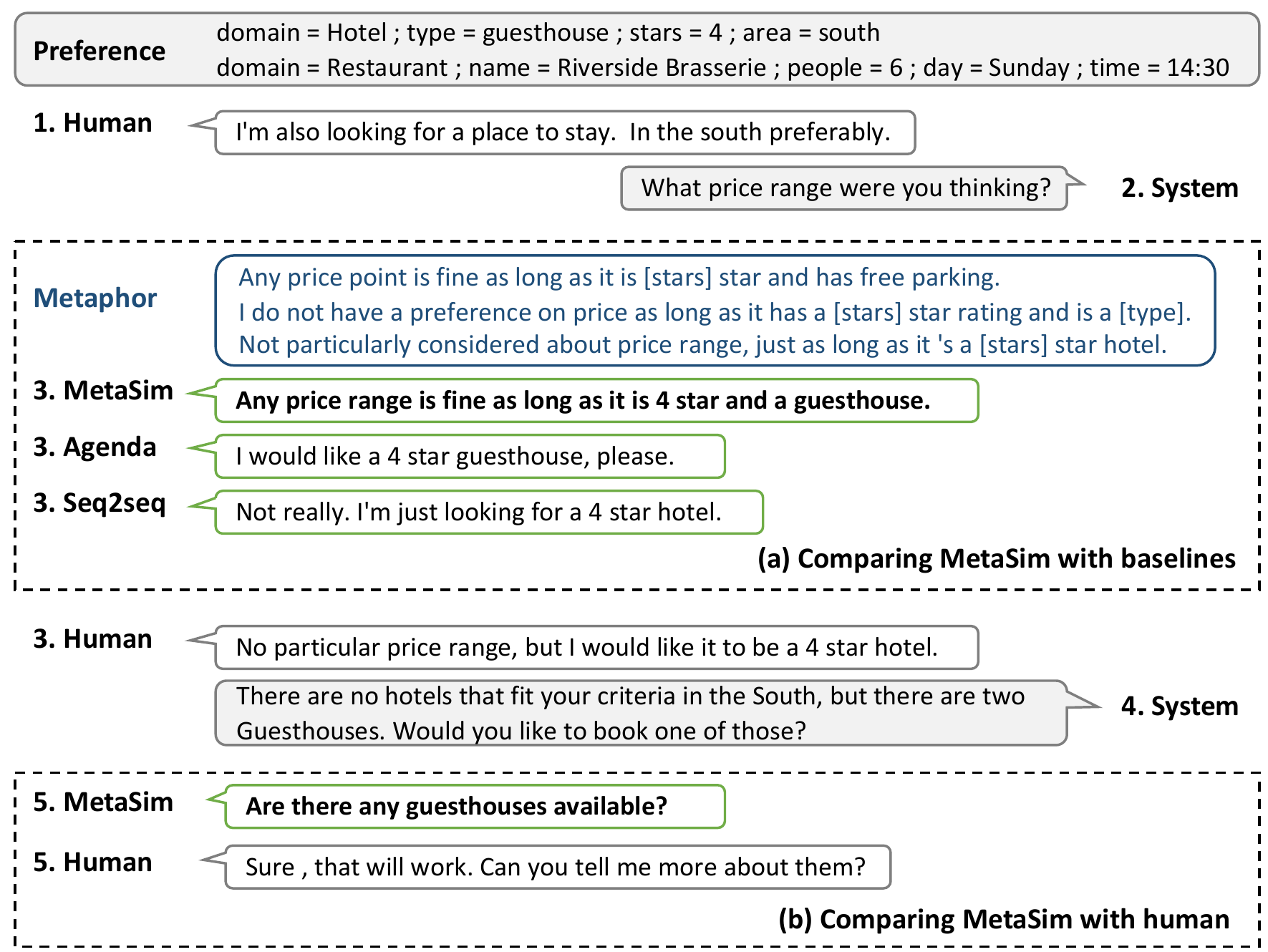}
 \caption{An example dialogue with responses generated by MetaSim, Agenda, and Seq2seq. In (a), we compare the predictions of MetaSim and baselines for turn 3. In (b), we compare the predictions of MetaSim and Human for turn 5.
 }
 \label{fig:cases}
\end{figure}

\begin{figure}[t]
 \centering
 \includegraphics[width=0.9\columnwidth]{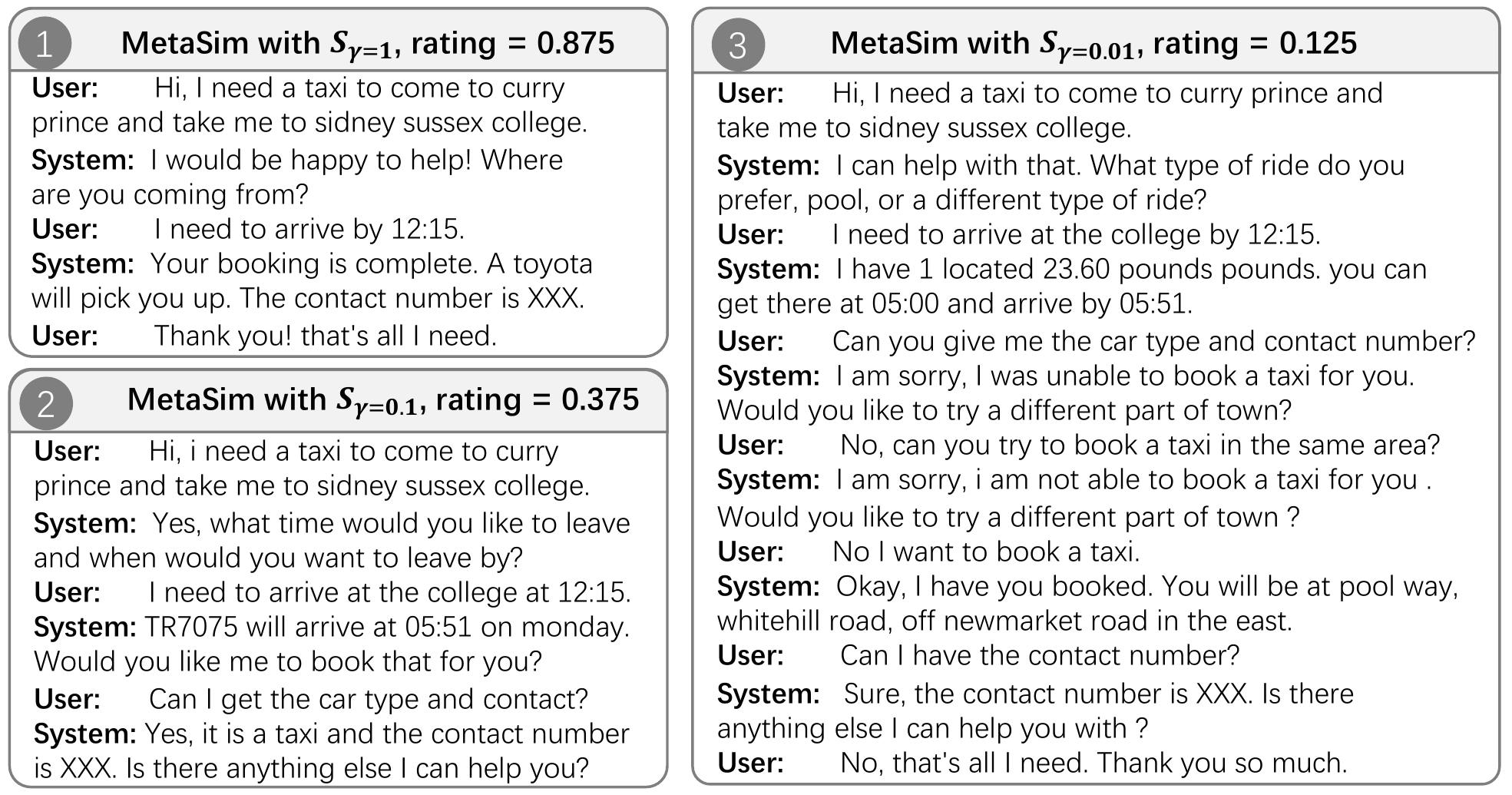}
 \caption{Simulated dialogues of MetaSim with Domain tester, as well as ratings by the simulator.}
 \label{fig:cases2}
\end{figure}

\subsection{Case studies}
\label{sec:sec:case}
We display examples of responses predicted by MetaSim, Agenda, Seq2seq, and a human annotator (all given the same user preference) for the same dialogue in Figure~\ref{fig:cases}. 
In Figure~\ref{fig:cases} (a), we see that the response generated by MetaSim is more natural and coherent to the dialogue than the Agenda model. 
In Figure~\ref{fig:cases} (b), we see a mistake by MetaSim, where human asks further information about Guesthouses while MetaSim fails to predict the action.

In Figure~\ref{fig:cases2}, we show examples of simulated dialogues of MetaSim with three variants defined by the Recommender tester, as well as the rating of the systems. 
These variants have sensible capabilities, as witnessed by these examples, i.e., $S_{\gamma=1}$ successfully books a taxi that meets the user's requirements, $S_{\gamma=0.1}$ completes a booking, but the booking time is wrong, and $S_{\gamma=0.01}$ fails in this task. MetaSim is able to distinguish between the three simulated dialogues with clearly different scores.

The two case studies illustrate that the MetaSim is capable of generating natural and coherent responses and of distinguishing between systems of different quality.


\section{Conclusions}
\label{section:conclusion}

In this paper, we have studied the problem of evaluating task-oriented dialogue systems using user simulators.
Based on the identified challenges when employing user simulators in evaluating \acp{TDS}, i.e., the lack of automatic methods to evaluate the user simulators, we have proposed a solution target at improving the realism of the current user simulator for better evaluating \ac{TDS}.
Specifically, we have proposed the metaphorical user simulator (MetaSim) for end-to-end TDS evaluation and a tester-based evaluation
framework for evaluating the evaluation capability of the simulators.

We have conducted extensive experiments on three benchmark \ac{TDS} datasets, i.e., MultiWOZ, ReDial, and JDDC.
Our experiments have shown that the proposed simulator MetaSim achieves better consistency with manual evaluation on those three datasets and that it is able generate more realistic dialogues than the Agenda-based simulator and a Seq2seq model.
We also found that system variants defined by different settings of the tester display meaningful performance differences. 

A broader lessons that we arrive at through our study is as follows.
Realism is important when constructing user simulators for evaluating dialogue systems.
In this work, we have taken important steps towards more realistic user simulators, by constructing mental models.
User simulators allowing for a more realistic assessment can reduce the reliance on human effort in the loop of evaluation.
Moreover, a test-based evaluation framework reduces the dependence on third-party agents, and can assess user simulators appropriately.

Recall that in Sect. \ref{sec:sec:policy}, we predict the user action and satisfaction jointly.
But the combination of user simulation and satisfaction assessment is still preliminary as we simply perform a sequence prediction and add the results of both (see Sect. \ref{sec:sec:inference}).
Also, our simulator design with multiple modules is computationally intensive and thus limits the inference efficiency. 
\sww{Moreover, our research on establishing mental models is still in its early stages, with limited evaluation of its effectiveness in multi-domain and cross-task performance.}

In future work, we would like to explore more effective combination of user simulation and satisfaction assessment.
In addition, we would like to improve tester-based evaluation strategies and simulator design so that it can be more efficient,
for example, reducing the number of interactions between tester and simulators and employing end-to-end modeling~\citep{Lei2018SequicityST}.
Moreover, we would like to 
extend the tester framework to support a broader range of simulator evaluation demands, such as conversational question answering and knowledge-grounded dialogue.

\section*{Data and resources}
This paper only made use of publicly available datasets. 
The code used to produce the results in this paper is available at \url{http://github.com/sunnweiwei/MetaSim}.
The code of tester-based framework is available at \url{https://github.com/Superbooming/simtester}.

\begin{acks}
This work was supported by the National Key R\&D Program of China with grant No. 2020YFB1406704,
and by the Hybrid Intelligence Center, a 10-year program funded by the Dutch Ministry of Education, Culture and Science through the Netherlands Organisation for Scientific Research, \url{https://hybrid-intelligence-centre.nl}.
All content represents the opinion of the authors, which is not necessarily shared or endorsed by their respective employers and/or sponsors.
\end{acks}

\bibliographystyle{ACM-Reference-Format}
\bibliography{references}

\end{document}